\documentclass[a4paper,fleqn,authoryear]{cas-sc}
\usepackage[authoryear]{natbib}
\usepackage{graphicx} 
\usepackage{adjustbox}
\usepackage{booktabs} 
\usepackage{xcolor}
\usepackage{soul} 
\sethlcolor{yellow}
\newcommand{\hlyellow}[1]{\sethlcolor{yellow}\hl{#1}} 
\newcommand{\hlblue}[1]{\textcolor{blue}{#1}}
\def\tsc#1{\csdef{#1}{\textsc{\lowercase{#1}}\xspace}}
\tsc{WGM}
\tsc{QE}
\tsc{EP}
\tsc{PMS}
\tsc{BEC}
\tsc{DE}

\begin{document}
\let\WriteBookmarks\relax
\def\floatpagepagefraction{1}
\def\textpagefraction{.001}

\shorttitle{Evaluation of Reasoning LLMs for Dialogue Summarization}

\shortauthors{Keyan Jin et~al.}

\title [mode = title]{Reasoning or Not? A Comprehensive Evaluation of Reasoning LLMs for Dialogue Summarization}                      

\author[1,4]{Keyan Jin}
\author[1]{Yapeng Wang}
\author[2]{Leonel Santos}
\author[3]{Tao Fang}
\author[1]{Xu Yang}
\author[1]{Sio Kei Im}
\author[4]{Hugo Gonçalo Oliveira}

\affiliation[1]{organization={Faculty of Applied Sciences, Macao Polytechnic University},
                city={Macao},
                country={China}}
\affiliation[2]{organization={Computer Science and Communication Research Center, School of Technology and Management, Polytechnic of Leiria},
                city={Leiria},
                country={Portugal}}
                
\affiliation[3]{organization={Institute of International Language Services Studies, Macau Millennium College},
                city={Macao},
                country={China}}
                
\affiliation[4]{organization={CISUC, Department of Informatics Engineering, University of Coimbra},
                city={Coimbra},
                country={Portugal}}

\begin{abstract}
Dialogue summarization is a challenging task with significant practical value in customer service, meeting analysis, and conversational AI. 
Although large language models (LLMs) have achieved substantial progress in summarization tasks, the performance of step-by-step reasoning architectures—specifically Long Chain-of-Thought (CoT) implementations such as OpenAI-o1 and DeepSeek-R1—remains unexplored for dialogue scenarios requiring concurrent abstraction and conciseness. 
In this work, we present the first comprehensive and systematic evaluation of state-of-the-art reasoning LLMs and non-reasoning LLMs across three major paradigms—generic, role-oriented, and query-oriented dialogue summarization. Our study spans diverse languages, domains, and summary lengths, leveraging strong benchmarks (SAMSum, DialogSum, CSDS, and QMSum) and advanced evaluation protocols that include both LLM-based automatic metrics and human-inspired criteria. Contrary to trends in other reasoning-intensive tasks, our findings show that explicit stepwise reasoning does not consistently improve dialogue summarization quality. Instead, reasoning LLMs are often prone to verbosity, factual inconsistencies, and less concise summaries compared to their non-reasoning counterparts. Through scenario-specific analyses and detailed case studies, we further identify when and why explicit reasoning may fail to benefit—or even hinder—summarization in complex dialogue contexts. Our work provides new insights into the limitations of current reasoning LLMs and highlights the need for targeted modeling and evaluation strategies for real-world dialogue summarization.
\end{abstract}



\maketitle
\section{Introduction}

Dialogue summarization is a critical natural language processing task that supports numerous practical applications, such as customer service, meeting analysis, and conversational AI assistants. Unlike traditional document summarization, dialogue summarization must handle unique challenges, including multi-party interactions, fragmented utterances, ambiguous references, and frequent topic shifts. Additionally, effective summarization can facilitate automated meeting documentation, collaborative decision-making, and efficient information retrieval from dialogue records. Early advances relied primarily on extractive methods that selected key sentences based on simple heuristics like TF–IDF or word frequency~\citep{marcu_discourse_1997}, before evolving to neural approaches such as Seq2Seq and Pointer-Generator networks, which enabled more fluent abstractive summaries~\citep{rush_neural_2015, see_get_2017}. Subsequently, significant breakthroughs were achieved by adapting Transformer-based neural architectures to conversational settings~\citep{lewis_bart_2019, liang_towards_2022, jin_span_2025}. Large language models (LLMs) have achieved remarkable results across a wide variety of natural language processing tasks, including text classification, sentiment analysis, question answering, and translation, demonstrating strong generalization capabilities and state-of-the-art performance~\citep{brown_language_2020}. In particular, reasoning LLMs, such as OpenAI-o1, DeepSeek-R1, and QwQ-32B, have exhibited notable advantages in tasks requiring complex reasoning, such as mathematical problem solving, logical inference, and machine translation~\citep{chen_evaluating_2025, ye_how_2025}. These successes naturally prompt further exploration into their applicability within dialogue summarization.

\begin{figure*}[t]
    \centering
    \includegraphics[width=\textwidth]{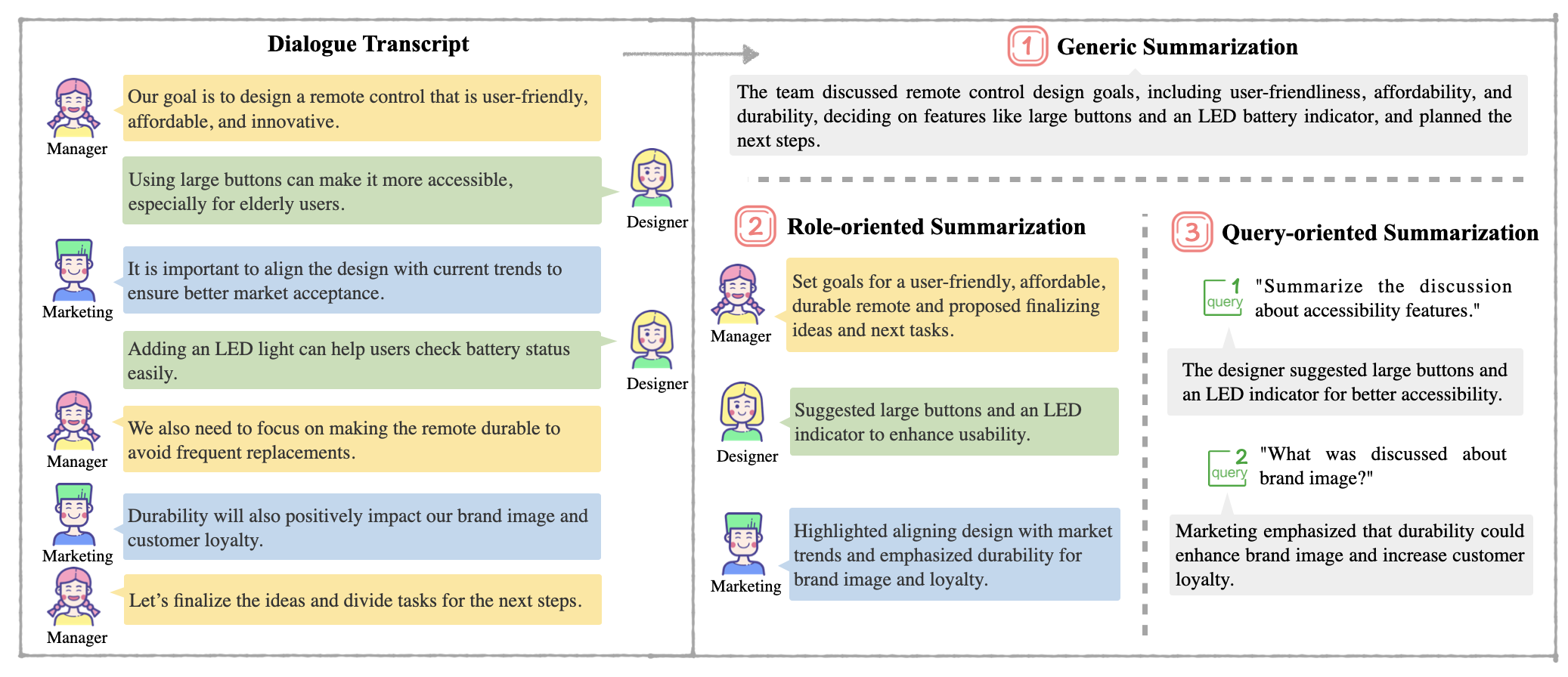}
    \caption{An illustrative example of three dialogue summarization tasks explored in this study: Generic Summarization, Role-oriented Summarization, and Query-oriented Summarization.}
    \label{task_example}
\end{figure*}

Dialogue summarization encompasses multiple distinct paradigms, each reflecting real-world scenarios that vary significantly in language, domain, dialogue length, and user intent. As illustrated in Figure~\ref{task_example}, the three widely recognized paradigms include generic summarization, which involves summarizing the entire dialogue; role-oriented summarization, focusing specifically on the perspectives of distinct dialogue participants; and query-oriented summarization, aiming to satisfy particular information requests. Despite considerable progress, existing research lacks a systematic and comprehensive evaluation of reasoning-based LLMs compared to traditional LLMs across diverse dialogue summarization scenarios. In particular, there is currently no multi-dimensional evaluation that examines their comparative performance across different summarization paradigms, languages, domains, and dialogue lengths.

In this work, we present a systematic, scenario-driven evaluation of reasoning and traditional LLMs for dialogue summarization. Our experiments cover three core paradigms—generic, role-oriented, and query-oriented summarization—across diverse benchmarks including SAMSum~\citep{gliwa2019samsum}, DialogSum~\citep{chen2021dialogsum}, CSDS~\citep{lin2021csds}, and QMSum~\citep{zhong2021qmsum}.  We comprehensively evaluate leading reasoning models (OpenAI-o1, DeepSeek-R1, QwQ-32B) and their non-reasoning counterparts across different languages, domains, and dialogue lengths. To ensure a fair and consistent comparison across models, our evaluation exclusively utilizes prompting, without any task-specific fine-tuning. By combining LLM-based automatic evaluation and human-inspired fine-grained metrics, we assess summary quality from multiple perspectives and investigate how explicit reasoning processes shape model performance. Our analysis is further enriched by detailed case studies that highlight the  limitations of reasoning LLMs in real-world dialogue summarization. Our main contributions are as follows:
\begin{itemize}
    \item We establish a unified framework for dialogue summarization, covering generic, role-oriented, and query-oriented paradigms, and provide clear definitions and scenario mappings for each.
    \item We present the first large-scale, systematic evaluation of leading reasoning-oriented LLMs and their non-reasoning counterparts on widely adopted dialogue summarization datasets, reporting results across multiple languages, domains, and summary types.
    \item We identify open challenges in automatic and LLM-based evaluation, and offer practical insights for selecting and developing models tailored to complex dialogue summarization needs.
\end{itemize}

The rest of the paper is organized as follows. Section 2 discusses related work on dialogue summarization and the evaluation of reasoning-oriented large language models. Section 3 describes our research methodology, including the problem definition, evaluation framework, datasets, and prompt design. In Section 4, we present experimental results covering generic, role-oriented, and query-oriented summarization tasks. Section 5 provides a detailed analysis of the intrinsic quality of summaries, comparative evaluations using large language models as judges, an in-depth analysis of reasoning processes, and a case study. Finally, Section 6 concludes the paper and discusses future research directions.

\section{Related Work}

In this section, we review recent advancements in dialogue summarization with LLMs, including their methodological innovations, emerging evaluation approaches, and existing challenges. We further discuss recent efforts to systematically evaluate reasoning LLMs, emphasizing their capabilities and limitations across complex reasoning tasks.

\subsection{Dialogue Summarization with LLMs}

Recent progress in dialogue summarization has been driven by the advent of LLMs, instruction tuning, and advanced prompt engineering. Early studies focused on fine-tuning smaller models or exploiting heuristic approaches, but instruction-tuned LLMs have shown remarkable ability in both generic and role-oriented dialogue summarization. For example, Baichuan2-Sum demonstrates that role-specific instructions can significantly improve summary quality for multiple roles within a dialogue~\citep{xiao2024baichuan2}. Multi-stage pre-training and mutual reinforcement between synthesis and summarization can further enhance LLM adaptability across diverse scenarios~\citep{zhou_multi-stage_2023,lu_mutual_2025}. Prompt optimization, such as prompt scoring and in-context example selection, has also proven critical for controlling summary content, style, and focus, with studies showing that both prompt format and demonstration quality can greatly affect summarization performance~\citep{tang_-context_2023,block_summary_2023,okadome_prompt_2024}. For cross-lingual and scenario adaptation, zero-shot and few-shot LLMs are now competitive with supervised approaches, but remain challenged by verbosity and cultural adaptation~\citep{wang_zero-shot_2023}.

A key challenge for LLM-based dialogue summarization lies in evaluating and improving factual consistency, faithfulness, and overall summary quality. Recent research underscores that traditional summarization metrics, particularly ROUGE, are inadequate for fully capturing essential quality dimensions like consistency and relevance in dialogue summarization tasks~\citep{gao_dialsummeval_2022}. As a result, a wide range of new evaluation frameworks and benchmarks have emerged. Fine-grained and multi-dimensional evaluation datasets—such as FineSurE~\citep{song_finesure_2024} and UniSumEval~\citep{lee_unisumeval_2024}—enable the systematic assessment of faithfulness, completeness, conciseness, and hallucination at the sentence or span level. Laban et al.~\citep{laban_summedits_2023} propose SummEdits, which directly tests models' ability to reason about facts in summaries. TofuEval further highlights that hallucinations remain a persistent issue even for advanced LLMs~\citep{tang_tofueval_2024}, and that LLM-based evaluators are not always superior to specialized factuality metrics. Newer error taxonomies, such as contextual or circumstantial inference~\citep{ramprasad_analyzing_2024}, reveal that LLMs often generate plausible but unsupported content, making automated detection difficult. Studies on guideline-following~\citep{zhou_can_2025}, extractive+abstractive hybrid models~\citep{zhang_extractive_2023,mishra_llm_2023}, and mixture-of-experts strategies~\citep{tian_dialogue_2024} further enrich the landscape of summary evaluation and model design.

Finally, beyond mainstream English conversation, dialogue summarization research is expanding to more complex and realistic domains. Multi-domain and multi-scenario benchmarks test LLMs' robustness across varied dialogue styles, speaker structures, and background knowledge requirements~\citep{zhou_multi-stage_2023}. Methods such as recursive summarization for long-term dialogue memory~\citep{wang_recursively_2025} and factual knowledge distillation from LLMs to smaller models~\citep{zhu_factual_2025} point to the growing ecosystem of hybrid, scalable, and adaptive approaches in the field. Collectively, these advances signal a paradigm shift toward LLM-powered dialogue summarization—supported by a rapidly evolving set of evaluation tools and practical solutions for controlling, scaling, and verifying model outputs in real-world applications.

\subsection{Evaluation of Reasoning LLMs}

Recent years have seen a surge of interest in reasoning LLMs, exemplified by OpenAI’s o1, DeepSeek-R1, QwQ, and the new Qwen3 series~\citep{deepseek-ai_deepseek-r1_2025, yang_qwen3_2025}. These models are specifically optimized to emulate human-like multi-step reasoning, often leveraging reinforcement learning or chain-of-thought (CoT) training paradigms to improve their performance on tasks involving complex logic, problem-solving, and decision-making~\citep{yang_qwen3_2025}. Recent works have demonstrated that such “reasoning LLMs” can achieve competitive or even state-of-the-art results on benchmarks requiring planning, mathematical reasoning, commonsense, or domain adaptation, often surpassing traditional LLMs in semantically demanding scenarios~\citep{ye_how_2025, valmeekam_systematic_2025}. However, it has also been observed that these models come with trade-offs: inference is typically slower and more costly, and models may suffer from issues such as overthinking, redundant reasoning, and unstable output length or quality~\citep{hashemi_dnr_2025, zeng_revisiting_2025}. For example, it has been shown that o1-like models often allocate excessive computational resources to simple problems, yielding limited benefits and resulting in inefficient scaling during inference~\citep{chen_not_2025}. Other studies have noted that longer reasoning chains do not always improve accuracy and may actually increase the likelihood of self-contradictions and unnecessary self-revisions~\citep{zeng_revisiting_2025}.

The evaluation of reasoning LLMs has grown increasingly systematic, with dedicated benchmarks for reflection, planning, factual reasoning, and human-like social cognition. For instance, the LR$^2$Bench was introduced to assess long-chain reflective reasoning and constraint satisfaction, showing that even the most advanced reasoning models still struggle on compositional logic and multi-step constraint problems, leaving substantial room for improvement~\citep{chen_lr2bench_2025}. Comparative studies have revealed that models like DeepSeek-R1, OpenAI-o1, and QwQ are effective on complex translation, mathematical, or planning tasks, but their performance can degrade due to over-reasoning and “rambling” in certain languages or domains~\citep{chen_evaluating_2025, ye_how_2025, valmeekam_systematic_2025}. Overthinking and excessive token generation remain open issues, as shown by DNR Bench, which probes whether models can recognize when reasoning is unnecessary and avoid generating superfluous content~\citep{hashemi_dnr_2025}. 

\section{Methodology}

In this section, we formally define the three dialogue summarization tasks explored in this study and describe our experimental framework, including datasets, prompts, and evaluation metrics.

\subsection{Problem Formulation}
\begin{figure*}[t]
    \centering
    \includegraphics[width=\textwidth]{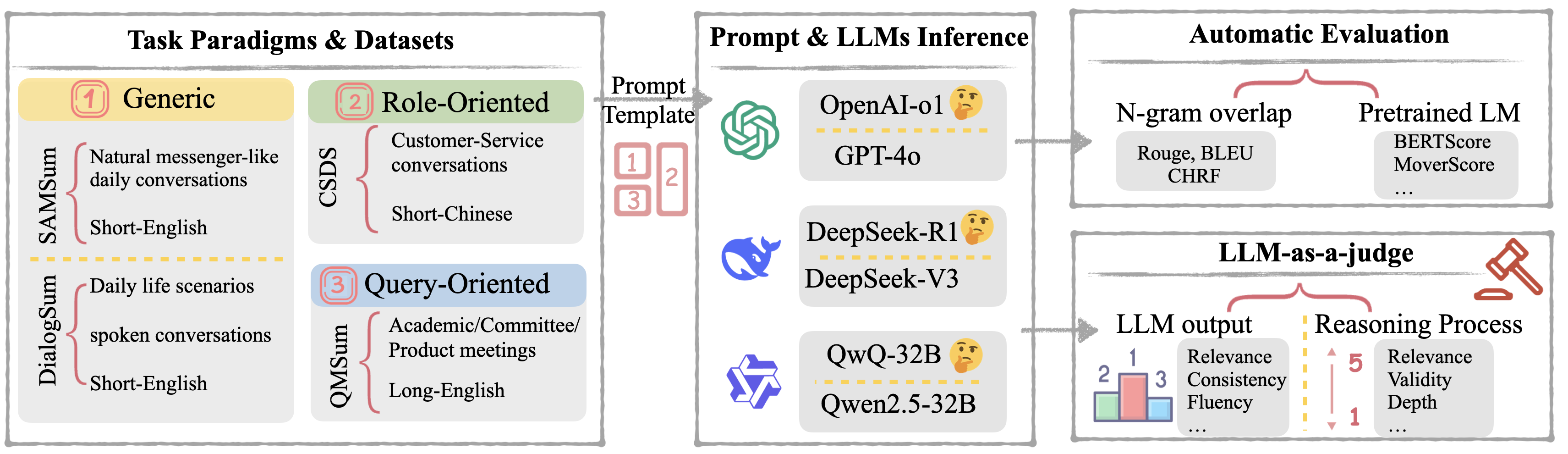}
    \caption{Overview of our experimental framework. We evaluate multiple reasoning and non-reasoning LLMs across three dialogue summarization paradigms using diverse datasets.}
    \label{framework}
\end{figure*}
In this study, we explore dialogue summarization tasks under three distinct scenarios: generic dialogue summarization, role-oriented dialogue summarization, and query-based dialogue summarization.
Formally, given a dialogue \(D\) composed of a sequence of utterances, each utterance \(u_i\) is associated with a speaker role \(s_i\). Thus, a dialogue with \(N\) utterances and \(M\) unique speaker roles can be denoted as:
\begin{equation}
D = \{(u_1,s_1), (u_2,s_2), \dots, (u_N,s_N)\}, \quad s_i \in \{s_1,\dots,s_M\}
\end{equation}

The generic dialogue summarization task aims to generate a concise summary \(Y^{final}\) that covers the essential content of the entire dialogue \(D\). This process can be modeled as a sequence-to-sequence generation task, formally defined as:
\begin{equation}
    Y^{final} = \arg\max_{Y} P(Y|D)
\end{equation}

In role-oriented dialogue summarization, we focus on generating summaries from the perspectives of specific speaker roles within the dialogue. Given the dialogue \(D\) and a set of predefined roles \(S=\{s_1,\dots,s_M\}\), the task is to produce distinct summaries \(Y^{s_j}\) tailored to each role \(s_j\):
\begin{equation}
    Y^{s_j} = \arg\max_{Y} P(Y|D, s_j), \quad s_j \in S
\end{equation}

Moreover, dialogues often involve multiple topics or decision points, necessitating the query-oriented dialogue summarization. Given a query \(Q\) composed of words \((w_1, w_2, \dots, w_{|Q|})\), the task produces a targeted summary \(Y^Q\) responsive to the query:
\begin{equation}
    Y^{Q} = \arg\max_{Y} P(Y|D, Q)
\end{equation}

Queries can be categorized into general queries (\(Q_{gen}\)), requesting overall dialogue summaries, and specific queries (\(Q_{spec}\)), focusing on particular topics or roles. Thus, query-based summarization tasks are explicitly defined as:
\begin{equation}
    Y^{gen} = \arg\max_{Y} P(Y|D, Q_{gen}), \quad
    Y^{spec} = \arg\max_{Y} P(Y|D, Q_{spec})
\end{equation}

\subsection{Evaluation Framework}

In this subsection, we present our evaluation framework designed to systematically assess and compare reasoning and non-reasoning LLMs. We introduce the overall structure of our evaluation process, the benchmark datasets selected for different summarization tasks, our standardized prompt design strategies, and the detailed automatic metrics used for quantitative evaluation.

\subsubsection{Overall Framework }

We establish a unified evaluation framework to enable a rigorous, scenario-specific comparison of reasoning and non-reasoning LLMs for dialogue summarization. The overall design of our framework is illustrated in Figure~\ref{framework}, which outlines each stage of our experimental pipeline—from task paradigms and datasets to model inference, and finally to comprehensive evaluation.

Our framework begins by covering three representative summarization paradigms: generic, role-oriented, and query-oriented summarization. Each paradigm is paired with real-world benchmark datasets that reflect a range of dialogue types, domains, languages, and summary requirements. This setup allows us to test models under a wide spectrum of realistic and challenging scenarios.

For each task, we use standardized prompt templates to ensure fairness and reproducibility across both reasoning LLMs and their non-reasoning counterparts.  Detailed descriptions of prompt design and task-specific configurations are presented in a dedicated section later in the paper.

To thoroughly assess model performance, our evaluation combines both automatic and LLM-based human-aligned metrics. We report a range of widely used n-gram overlap metrics as well as pretrained language model-based scores. Beyond these surface-level measures, we incorporate LLM-as-a-judge protocols: large models are used as automatic evaluators to rank or rate generated summaries on key aspects such as relevance, consistency, fluency, and overall quality. For reasoning LLMs that output explicit stepwise traces, we further evaluate the quality of their reasoning process along targeted criteria.

\subsubsection{Datasets}

\begin{table*}[|t]
\centering
\caption{Statistics of datasets used in this work. Length is measured in average number of tokens.}
\footnotesize
\begin{adjustbox}{max width=\textwidth}
\renewcommand{\arraystretch}{1.2}
\begin{tabular}{l c l r r r r r l}
\Xhline{0.7pt}
\textbf{Datasets} & \textbf{Lang.} & \textbf{Description} & \textbf{Length} & \textbf{\#Turns} & \textbf{Length of Sum} & \textbf{\#Speakers} & \textbf{Dialogues} & \textbf{Sum Type} \\
\Xhline{0.7pt}
\textbf{SAMSum} & EN & Written messenger online & 95.51 & 11.25 & 23.12 & 2.36 & 819 & generic \\
\textbf{DIALOGSUM} & EN & Daily life spoken & 134.46 & 13.85 & 18.75 & 2 & 1500 & generic \\
\textbf{CSDS} & CN & Customer service & 387.10 & 25.10 & \makecell{Overall / User / Agent\\83.21 / 37.28 / 48.08} & 2 & 800 & generic / role based \\
\hline
\textbf{QMSum} & & & & & & & & \\
Academic & EN & Academic meetings & 13317.30 & 819.00 & 53.70 & 6.30 & 56 & generic / query based \\
Committee & EN & Committee meetings & 13761.90 & 207.70 & 80.50 & 34.10 & 72 & generic / query based \\
Product & EN & Product meetings & 6007.71 & 535.60 & 70.50 & 4.00 & 151 & generic / query based \\
ALL & EN & N/A & 9069.8 & 556.80 & 69.60 & 9.20 & 279 & generic / query based \\
\Xhline{0.7pt}
\end{tabular}
\end{adjustbox}
\label{tab:dataset_statistics}
\end{table*}

To comprehensively evaluate the summarization performance of reasoning-oriented o-1-like models and non-o1-like models, we select four representative datasets: \textbf{SAMSum}\footnote{\url{https://huggingface.co/datasets/Samsung/samsum}} \citep{gliwa2019samsum}, \textbf{DIALOGSUM}\footnote{\url{https://github.com/cylnlp/DialogSum}} \citep{chen2021dialogsum}, \textbf{CSDS}\footnote{\url{https://github.com/xiaolinAndy/CSDS}} \citep{lin2021csds}, and \textbf{QMSum}\footnote{\url{https://github.com/Yale-LILY/QMSum}} \citep{zhong2021qmsum}. Each dataset possesses unique characteristics, enabling evaluation of summarization capabilities under diverse dialogue scenarios and summarization tasks. The detailed statistics of these datasets are summarized in Table~\ref{tab:dataset_statistics}.

\textbf{SAMSum} is widely recognized for its manually annotated, high-quality messenger-style dialogues, primarily reflecting informal online communication scenarios. It is extensively used as a benchmark dataset in dialogue summarization research, allowing us to directly compare and assess the summarization performance of various models within informal conversational contexts.

\textbf{DIALOGSUM} is distinctive due to its extensive collection of spoken dialogues sourced from public corpora and an English speaking practice website. The dialogues represent typical daily-life scenarios such as education, employment, healthcare, shopping, and leisure activities. The dataset uniquely emphasizes observer-perspective summarization, making it highly suitable for evaluating models' abilities to generalize summarization tasks across various everyday conversational settings.

\textbf{CSDS} specifically addresses customer service dialogues in Chinese, offering distinctive features such as role-oriented summarization alongside general summaries. This characteristic allows the evaluation of summarization methods in handling complex dialogues involving distinct speaker perspectives (customers and service agents) and clearly structured topics. Such features position CSDS as a critical dataset for assessing models' capabilities in nuanced, topic-centric summarization tasks.

\textbf{QMSum} is uniquely designed for the summarization of lengthy, multi-participant meetings across academic, committee, and product development contexts. Distinctively employing a query-based summarization paradigm, QMSum challenges models to selectively generate summaries based on targeted queries. This approach provides an effective evaluation scenario to test summarization accuracy and relevance within complex and extensive dialogues.

\subsubsection{Prompt Design}
\begin{figure}[t]
    \centering
    \includegraphics[width=0.8\linewidth]{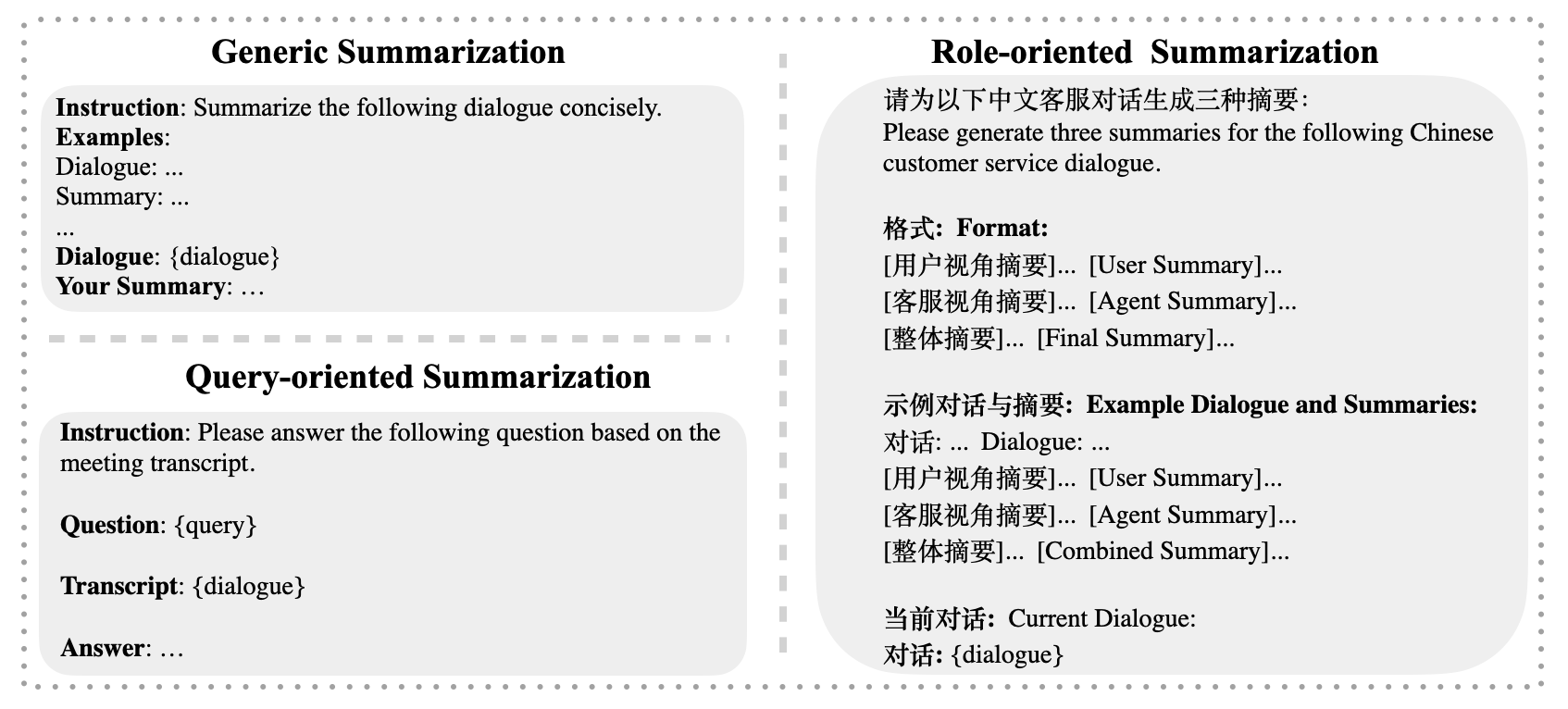} 
    \caption{Prompt templates for generic, role-oriented, and query-based dialogue summarization tasks used in our experiments. Note that for role-oriented summarization, since the evaluated dataset is in Chinese, we directly use Chinese prompts as shown. The English translations in the figure are provided for clarity.}
    \label{prompt_template}
\end{figure}

In terms of prompt construction, we carefully designed task-specific prompts for each of the three summarization scenarios: generic summarization, role-oriented summarization, and query-based summarization. The detailed prompt templates for each task are illustrated in Figure \ref{prompt_template}. 

\subsubsection{Outcome Evaluation}

We comprehensively evaluate the summarization quality of reasoning and non-reasoning models using various automatic evaluation metrics. These metrics are broadly categorized into two groups: traditional n-gram overlap-based metrics and pretrained language model-based metrics. We describe each group of metrics in detail below.

\paragraph{Traditional N-gram Overlap-based Metrics}  
These metrics evaluate the lexical similarity between the generated summaries \(Y\) and reference summaries \(Y^*\) based on n-gram overlaps:

\textbf{ROUGE}\footnote{\url{https://github.com/Diego999/py-rouge}} \citep{lin2004rouge}: Specifically, we use ROUGE-1, ROUGE-2, and ROUGE-L, capturing unigram, bigram, and longest common subsequence overlaps, respectively. ROUGE-N is computed as:
    \begin{equation}
        \text{ROUGE-N} = \frac{\sum_{S\in Y^*}\sum_{\text{gram}_n\in S} \text{Count}_{match}(\text{gram}_n)}{\sum_{S\in Y^*}\sum_{\text{gram}_n\in S} \text{Count}(\text{gram}_n)}
    \end{equation}
    
\textbf{BLEU}\footnote{\url{https://www.nltk.org/_modules/nltk/translate/bleu_score.html}} 
 \citep{papineni2002bleu}: BLEU evaluates precision by calculating the geometric mean of modified n-gram precisions (usually up to four-grams), with a brevity penalty (BP) to discourage overly short summaries:
    \begin{equation}
        \text{BLEU} = BP \cdot \exp\left(\sum_{n=1}^{N} w_n \log p_n\right)
    \end{equation}
    
\textbf{CHRF}\footnote{\url{https://github.com/mjpost/sacrebleu}} \citep{popovic2015chrf}: CHRF calculates the harmonic mean (F-score) based on precision and recall at the character n-gram level, defined formally as:
    \begin{equation}
        \text{CHRF} = (1+\beta^2)\frac{\text{chrP}\cdot \text{chrR}}{\beta^2 \cdot \text{chrP} + \text{chrR}}
    \end{equation}
    Typically, \(\beta=2\) emphasizes recall.

\paragraph{Pretrained Language Model-based Metrics}  
These metrics leverage pretrained language models to evaluate semantic similarity beyond simple lexical overlaps, capturing deeper semantic alignments between generated and reference summaries:

\textbf{BERTScore}\footnote{\url{https://github.com/Tiiiger/bert_score}. We use \textit{roberta-large} for English and \textit{bert-base-chinese} for Chinese.}~\citep{Zhang*2020BERTScore}: Utilizes contextual embeddings derived from pretrained BERT models to measure semantic similarity between candidate summary tokens \(\hat{x}\) and reference summary tokens \(x\). Specifically, precision (\(P_{\text{BERT}}\)), recall (\(R_{\text{BERT}}\)), and F1-score (\(F_{\text{BERT}}\)) are computed as:
\begin{equation}
    R_{\text{BERT}} = \frac{1}{|x|}\sum_{x_i\in x}\max_{\hat{x}_j\in \hat{x}}x_i^\top \hat{x}_j,\quad
    P_{\text{BERT}} = \frac{1}{|\hat{x}|}\sum_{\hat{x}_j\in \hat{x}}\max_{x_i\in x}x_i^\top \hat{x}_j,\quad
    F_{\text{BERT}} = 2\frac{P_{\text{BERT}}\cdot R_{\text{BERT}}}{P_{\text{BERT}}+R_{\text{BERT}}}.
\end{equation}

\textbf{MoverScore}\footnote{\url{https://github.com/AIPHES/emnlp19-moverscore}. We use \textit{distilbert-base-uncased} for English and \textit{bert-base-chinese} for Chinese.} \citep{zhao2019moverscore}: Employs optimal transport (Earth Mover's Distance) to calculate semantic distances between embeddings from generated and reference summaries. The metric finds the minimum "cost" of aligning candidate and reference embeddings, thus capturing nuanced semantic similarities:
    \begin{equation}
        \text{MoverScore} = 1 - \text{EMD}(E(Y), E(Y^*))
    \end{equation}
    where \(E(\cdot)\) denotes the contextual embeddings of the summary.
    
\textbf{BARTScore}\footnote{\url{https://github.com/neulab/BARTScore}. We use \textit{bart-large-cnn} for English and \textit{bart-base-chinese} for Chinese.} \citep{yuan2021bartscore}: Evaluates summaries using pretrained BART sequence-to-sequence models, measuring the log-likelihood scores across different directions including source-to-hypothesis (s-h), reference-to-hypothesis (r-h), and hypothesis-to-reference (h-r). The general form is:
    \begin{equation}
        \text{BARTScore} = \frac{1}{m}\sum_{i=1}^{m}\log P(y_i|y_{<i}, x;\theta_{\text{BART}})
    \end{equation}

\textbf{COMET}\footnote{\url{https://huggingface.co/Unbabel/wmt22-comet-da}.We use \textit{wmt22-comet-da} for evaluation} \citep{rei2020comet}: Originally proposed for machine translation evaluation, COMET leverages multilingual pretrained language models (such as XLM-Roberta) to calculate semantic similarity scores. Given source text \(src\), hypothesis \(hyp\), and reference \(ref\), COMET predicts a quality score \(q\):
    \begin{equation}
        q = f_{\text{COMET}}(E(src), E(hyp), E(ref))
    \end{equation}

\section{Experiments}
In this section, we describe the selected models, experimental settings, and detailed results of our comprehensive evaluation across the three dialogue summarization paradigms.

\begin{figure}[t]
    \centering
    \includegraphics[width=\textwidth]{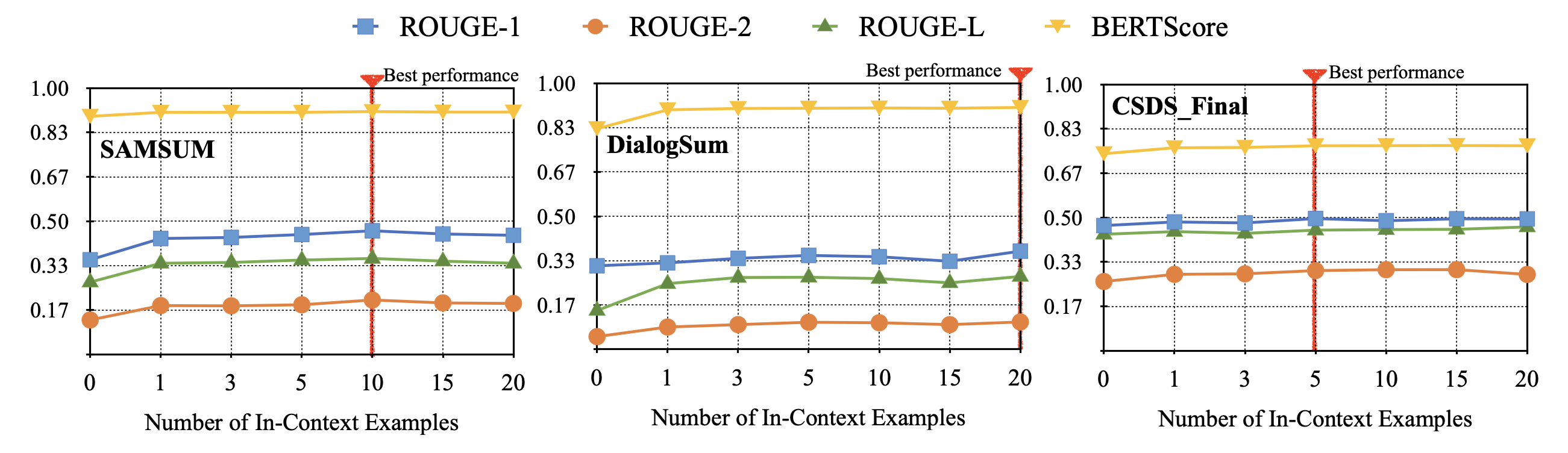}
    \caption{
    Performance comparison under varying numbers of in-context examples across different datasets using \texttt{gpt-4o}. 
    The best-performing number of in-context examples for each dataset is highlighted by a vertical red line and annotated as ``Best performance.'' 
    For \textbf{CSDS}, we report results based on the \textbf{Final} evaluation set. 
    Metrics include ROUGE-1, ROUGE-2, ROUGE-L, and BERTScore. 
    }
    \label{fig:icl_effect}
\end{figure}

\subsection{Model Selection}
To thoroughly evaluate dialogue summarization performance under different reasoning capabilities, we selected three reasoning LLMs: OpenAI-o1, DeepSeek-R1, and QwQ-32B. Each of these reasoning models is directly compared against a corresponding non-reasoning base model, ensuring fair and insightful evaluation.

\textbf{OpenAI-o1} is a reasoning-optimized model developed by OpenAI, designed to tackle complex tasks in science, coding, and mathematics. It employs a "think before answering" approach, generating internal chains of thought to enhance its problem-solving abilities. This model is compared with \textbf{GPT-4o}\footnote{We hypothesize that OpenAI-o1 and GPT-4o share the same foundational architecture; however, explicit confirmation has not been publicly disclosed.}.

\textbf{DeepSeek-R1}~\citep{deepseek-ai_deepseek-r1_2025} is a reasoning-specialized LLM developed by DeepSeek-AI, trained using large-scale reinforcement learning to solve complex reasoning tasks across domains such as math, code, and language. Its reasoning capabilities are further enhanced through supervised fine-tuning to improve readability and coherence. DeepSeek-R1 is compared with \textbf{DeepSeek-V3}, as they share the same underlying base architecture.

\textbf{QwQ-32B}~\citep{yang_qwen3_2025} is a medium-sized reasoning model from the Qwen series, developed by Alibaba's Qwen team. It leverages reinforcement learning techniques to enhance its reasoning capabilities, achieving competitive performance in mathematical reasoning, coding, and complex problem-solving tasks. QwQ-32B is evaluated against \textbf{Qwen-2.5-32B-Instruct}, its non-reasoning counterpart.
\subsection{Experimental Settings}
To evaluate the selected LLMs under realistic and reproducible conditions, all experiments were conducted via public APIs, ensuring consistent inference environments across models. Specifically, both OpenAI-o1 and GPT-4o were accessed using the official OpenAI API. For DeepSeek-R1, DeepSeek-V3, QwQ-32B, and Qwen-2.5-32B-Instruct, we utilized the APIs provided by the Aliyun Bailian platform\footnote{\url{https://bailian.console.aliyun.com/}}. 
For all models, We used the default parameters provided by the APIs for all models unless otherwise specified. 

\subsection{Experimental Results}

In this section, we first conduct preliminary ablation experiments to determine optimal experimental configurations. Then, we systematically evaluate the performance of reasoning and non-reasoning LLMs across three dialogue summarization scenarios. We analyze and compare their strengths and limitations using multiple automated evaluation metrics.

\begin{table*}[htbp]
\centering
\caption{Summarization performance of reasoning and non-reasoning LLMs on \textsc{SAMSUM} and \textsc{DialogSum} datasets using 14 automatic evaluation metrics. Best values are in bold.}
\footnotesize
\begin{adjustbox}{max width=\textwidth}
\renewcommand{\arraystretch}{1.2}
\begin{tabular}{l lcccccccccccccc}
\Xhline{0.7pt}
\textbf{} & \textbf{Model} & \textbf{R1} & \textbf{R2} & \textbf{R-L} & \textbf{BLEU} & \textbf{CHRF} 
& \multicolumn{3}{c}{\textbf{BERTScore}} & \textbf{Mover} 
& \multicolumn{3}{c}{\textbf{BARTScore}} & \textbf{COMET} & \textbf{Time(s)} \\
\cline{8-10} \cline{12-14}
& & & & & & & \textbf{P} & \textbf{R} & \textbf{F1} & & \textbf{S$\to$H} & \textbf{R$\to$H} & \textbf{H$\to$R} & & \\
\hline
\multirow{6}{*}{\rotatebox{90}{Samsum}} 
& \textbf{Reasoning LLMs} & & & & & & & & & & & & & & \\
& OpenAI-o1 & 45.06 & 18.54 & 35.33 & 7.83 & 41.26 & 89.58 & 91.88 & 90.70 & 26.07 & -2.17 & -2.69 & -2.24 & 72.96 & 5.49 \\
& DeepSeek-R1 & 43.90 & 17.50 & 34.30 & 7.64 & 41.07 & 89.16 & 91.80 & 90.45 & 24.55 & -2.40 & -2.99 & -2.25 & 71.98 & 22.06 \\
& QwQ-32B & 43.58 & 17.42 & 34.05 & 7.60 & 39.56 & 89.55 & 91.55 & 90.52 & 24.36 & -2.31 & -2.89 & -2.35 & 71.60 & 12.98 \\
& \textbf{Non-Reasoning LLMs} & & & & & & & & & & & & & & \\
& GPT-4o & 47.01 & 20.79 & 37.13 & \textbf{9.62} & 43.03 & \textbf{90.28} & 92.13 & \textbf{91.18} & \textbf{28.81} & -2.08 & -2.60 & \textbf{-2.21} & \textbf{74.09} & \textbf{1.32} \\
& DeepSeek-v3 & \textbf{47.21} & \textbf{20.94} & \textbf{37.67} & 9.61 & \textbf{43.07} & 90.16 & \textbf{92.15} & 91.13 & 28.44 & \textbf{-2.03} & \textbf{-2.55} & -2.22 & 74.03 & 3.28 \\
& Qwen2.5-32B & 44.88 & 19.24 & 35.45 & 8.75 & 41.73 & 89.69 & 91.96 & 90.79 & 26.02 & -2.17 & -2.73 & -2.24 & 73.11 & 2.77 \\
\hline
\multirow{6}{*}{\rotatebox{90}{DialogSum}} 
& \textbf{Reasoning LLMs} & & & & & & & & & & & & & & \\
& OpenAI-o1 & 37.99 & 13.25 & 29.67 & 5.33 & 42.27 & 88.66 & 91.80 & 90.19 & 14.03 & -2.05 & -2.70 & -2.41 & 70.12 & 9.83 \\
& DeepSeek-R1 & 37.59 & 12.69 & 29.26 & 5.31 & 41.62 & 88.63 & 91.52 & 90.04 & 13.65 & -2.40 & -2.97 & -2.44 & 69.28 & 48.06 \\
& QwQ-32B & 36.11 & 10.88 & 27.81 & 4.64 & 38.44 & 88.64 & 90.68 & 89.63 & 11.28 & -2.51 & -3.11 & -2.71 & 67.65 & 13.13 \\
& \textbf{Non-Reasoning LLMs} & & & & & & & & & & & & & & \\
& GPT-4o & 40.76 & 15.06 & 31.93 & 6.47 & 43.90 & 89.79 & 92.10 & \textbf{90.92} & 18.27 & -2.06 & -2.59 & -2.32 & 72.28 & 5.42 \\
& DeepSeek-v3 & \textbf{41.51} & \textbf{16.25} & \textbf{32.93} & \textbf{7.06} & \textbf{44.37} & \textbf{89.82} & \textbf{92.07} & \textbf{90.92} & \textbf{18.93} & \textbf{-1.92} & \textbf{-2.48} & \textbf{-2.36} & \textbf{72.40} & 8.81 \\
& Qwen2.5-32B & 39.17 & 13.79 & 30.78 & 5.84 & 42.41 & 89.41 & 91.90 & 90.62 & 16.51 & -2.17 & -2.69 & -2.39 & 71.15 & \textbf{3.64} \\
\Xhline{0.7pt}
\end{tabular}
\end{adjustbox}

\label{tab:final_samsum_dialogsum}
\end{table*}

\subsubsection{Preliminary Ablation Results}
To ensure the optimality of experimental outcomes, our study systematically conducted ablation experiments to investigate the selection of example quantities in few-shot learning. Specifically, we employed the GPT-4o API to perform In-Context Learning (ICL) ablation experiments \citep{dong2024survey}, across both general summarization and role-oriented summarization tasks. To comprehensively evaluate the impact of ICL example quantities on model performance, we configured varying numbers of example samples—namely, 1, 3, 5, 10, 15, and 20—and conducted a detailed analysis of the results. Performance evaluation was carried out using multiple automated metrics, including ROUGE-1, ROUGE-2, ROUGE-L, and BERTScore, to ensure the comprehensiveness and objectivity of the assessment.

The experimental results, illustrated in Figure \ref{fig:icl_effect}, reveal the performance trends across different ICL example quantity configurations on the validation sets of each dataset. We determined the optimal ICL example quantities for the SAMSum, DialogSum, and CSDS datasets to be 10, 20, and 5, respectively. This optimal configuration will be applied in all subsequent experiments to ensure the stability and consistency of model performance.

For the query-oriented summarization task (QMSum dataset), we noted that conference transcript texts are typically lengthy, often approaching or exceeding the token limits of the API. Consequently, we adopted a zero-shot prompting strategy. This approach not only effectively addressed the challenges of processing long texts but also significantly enhanced the feasibility and computational efficiency of dialogue summarization evaluation. 



\subsubsection{Evaluation of Generic Dialogue Summarization}

As summarized in Table~\ref{tab:final_samsum_dialogsum}, non-reasoning LLMs demonstrate a clear and consistent advantage over reasoning LLMs across both the SAMSum and DialogSum datasets. Non-reasoning LLMs consistently outperform reasoning LLMs across almost all automatic evaluation metrics.

Notably, DeepSeek-V3 and GPT-4o achieve the strongest overall performance on both datasets. For example, on SAMSum, DeepSeek-V3 obtains the highest ROUGE-1/2/L and CHRF scores, while GPT-4o achieves the best BERTScore (F1), MoverScore, and inference efficiency. Similar patterns are observed on DialogSum, with DeepSeek-V3 again leading most metrics, closely followed by GPT-4o. The performance gap is especially pronounced in metrics that reflect both surface overlap (ROUGE, BLEU, CHRF) and semantic similarity (BERTScore, COMET).

In contrast, reasoning LLMs consistently lag behind, both in terms of summary quality and inference speed. Despite being optimized for multi-step reasoning, these models fail to deliver notable advantages for generic dialogue summarization tasks, and in some cases, such as DeepSeek-R1, exhibit significantly slower inference.

These results clearly suggest that explicit reasoning instruction tuning does not generally benefit dialogue summarization tasks that emphasize brevity and direct extraction of core content. Nonetheless, we recognize a potential strength of reasoning LLMs not fully captured by conventional automatic metrics: their ability to explicitly outline decision-making processes or generate plausible explanations supporting summarization choices. Such reasoning transparency could be advantageous in scenarios requiring interpretability and trust, even though these benefits may not directly enhance measured summary quality.

\subsubsection{CSDS: Role-Oriented Chinese Dialogue Summarization}

Table~\ref{tab:csds_role_eval} reports the results for all models on the CSDS dataset, which focuses on Chinese customer service dialogues and provides reference summaries from user, agent, and overall perspectives. Across all roles, the reasoning LLMs consistently underperform compared to their respective base models on nearly all automatic metrics.

For instance, in the user summary setting, DeepSeek-V3 achieves the highest ROUGE score, while its reasoning variant DeepSeek-R1 scores notably lower on all these metrics. Similar trends are observed for GPT-4o versus OpenAI-o1 and for Qwen2.5-32B versus QwQ-32B. In the agent summary task, DeepSeek-V3 again leads with the best ROUGE and BERTScore values, and the gap between base and reasoning-tuned models remains substantial. The final (generic) summary results reinforce this observation: DeepSeek-V3 and other base models consistently achieve higher scores across ROUGE, BLEU, CHRF, and BERTScore than their reasoning counterparts.

These consistent patterns suggest that reasoning instruction tuning does not enhance performance for role-oriented abstractive summarization in Chinese dialogues. On the contrary, it often leads to reduced summary quality. One plausible explanation is that such tuning encourages models to engage in excessive reasoning or elaboration, which may detract from the concise and focused nature required for effective summarization. Consequently, general-purpose models, when equipped with appropriate prompting and in-context learning strategies, exhibit better adaptability and summarization quality across diverse dialogue scenarios and perspectives.

\begin{table*}[ht]
\centering
\caption{Summarization performance of reasoning LLMs and non-reasoning LLMs on \textsc{CSDS} dataset from three role perspectives. Best values are in bold.}
\footnotesize
\begin{adjustbox}{max width=\textwidth}
\renewcommand{\arraystretch}{1.2}
\begin{tabular}{l lcccccccccccccc}
\Xhline{0.7pt}
\textbf{Role} & \textbf{Model} & \textbf{R1} & \textbf{R2} & \textbf{R-L} & \textbf{BLEU} & \textbf{CHRF}
& \multicolumn{3}{c}{\textbf{BERTScore}} & \textbf{Mover}
& \multicolumn{3}{c}{\textbf{BARTScore}} & \textbf{COMET} \\
\cline{8-10} \cline{12-14}
& & & & & & & \textbf{P} & \textbf{R} & \textbf{F1} & & \textbf{S$\to$H} & \textbf{R$\to$H} & \textbf{H$\to$R} & \\
\hline

\multirow{9}{*}{\large User} 
& \textbf{Reasoning LLMs} & & & & & & & & & & & & & \\
& OpenAI-o1 & 49.75 & 30.16 & 45.06 & 16.16 & 17.17 & 75.86 & 78.06 & 76.77 & 51.88 & -4.36 & -3.67 & -3.44 & 82.93 \\
& DeepSeek-R1 & 51.50 & 32.23 & 47.02 & 17.55 & 18.31 & 76.29 & 78.16 & 77.02 & 50.17 & -4.40 & -3.65 & -3.38 & 82.83 \\
& QwQ-32B & 49.55 & 30.05 & 44.65 & 16.42 & 17.27 & 74.60 & 77.35 & 76.25 & 50.80 & -4.42 & -3.76 & -3.39 & 83.50 \\
& \textbf{Non-Reasoning LLMs} & & & & & & & & & & & & & \\
& GPT-4o & 52.53 & 33.61 & 48.39 & 18.73 & 18.11 & 77.95 & 78.07 & 77.82 & 52.32 & -4.39 & -3.55 & -3.42 & 83.43 \\
& DeepSeek-v3 & \textbf{54.40} & \textbf{36.58} & \textbf{50.74} & \textbf{20.84} & \textbf{19.83} & \textbf{79.88} & \textbf{78.78} & \textbf{79.13} & \textbf{54.34} & \textbf{-4.38} & \textbf{-3.43} & \textbf{-3.37} & \textbf{84.18} \\
& Qwen2.5-32B & 51.57 & 32.56 & 47.35 & 17.91 & 17.99 & 77.22 & 78.21 & 77.52 & 52.06 & -4.39 & -3.61 & -3.42 & 83.63 \\
\hline
\multirow{9}{*}{\large Agent}  
& \textbf{Reasoning LLMs} & & & & & & & & & & & & & \\
& OpenAI-o1 & 46.84 & 28.14 & 41.92 & 14.89 & 15.59 & 75.00 & 75.16 & 74.90 & 51.59 & -4.37 & -3.65 & -3.56 & 82.34 \\
& DeepSeek-R1 & 46.45 & 27.85 & 41.62 & 14.11 & 15.15 & 74.40 & 74.60 & 74.32 & 50.88 & -4.42 & -3.71 & -3.55 & 82.12 \\
& QWQ-32B & 45.83 & 26.61 & 40.57 & 13.56 & 14.71 & 73.53 & 74.94 & 74.08 & 51.12 & -4.41 & -3.77 & -3.57 & 82.61 \\
& \textbf{Non-Reasoning LLMs} & & & & & & & & & & & & & \\
& GPT-4o & 48.15 & 29.77 & 43.33 & 15.83 & 16.38 & 75.35 & 75.32 & 75.16 & 51.77 & -4.34 & -3.61 & -3.52 & 83.11 \\
& DeepSeek-v3 & \textbf{49.86} & \textbf{32.37} & \textbf{45.45} & \textbf{17.87} & \textbf{18.11} & \textbf{76.87} & \textbf{75.97} & \textbf{76.22} & \textbf{52.83} & \textbf{-4.30} & \textbf{-3.48} & \textbf{-3.46} & \textbf{83.39} \\
& Qwen2.5-32B & 48.01 & 29.66 & 43.42 & 16.17 & 16.75 & 75.12 & 75.73 & 75.26 & 52.54 & -4.33 & -3.62 & -3.51 & 83.59 \\
\hline
\multirow{9}{*}{\large Final} 
& \textbf{Reasoning LLMs} & & & & & & & & & & & & & \\
& OpenAI-o1 & 51.03 & 31.09 & 46.61 & 17.43 & 17.69 & 76.14 & 76.86 & 76.37 & 58.29 & -4.12 & -3.46 & -3.33 & 83.25 \\
& DeepSeek-R1 & 51.54 & 32.28 & 47.63 & 17.89 & 18.04 & 76.09 & 76.69 & 76.26 & 56.78 & -4.18 & -3.49 & -3.31 & 83.22 \\
& QwQ-32B & 49.87 & 29.98 & 45.41 & 16.68 & 17.16 & 74.63 & 76.68 & 75.52 & 57.42 & -4.17 & -3.55 & -3.33 & 83.50 \\
& \textbf{Non-Reasoning LLMs} & & & & & & & & & & & & & \\
& GPT-4o & 53.10 & 34.02 & 49.21 & 19.43 & 18.82 & 77.55 & 77.09 & 77.19 & 58.27 & -4.11 & -3.38 & -3.31 & 84.38 \\
& DeepSeek-v3 & \textbf{54.44} & \textbf{36.53} & \textbf{51.07} & \textbf{21.44} & \textbf{20.41} & \textbf{79.01} & \textbf{77.77} & \textbf{78.24} & \textbf{59.16} & \textbf{-4.08} & \textbf{-3.28} & \textbf{-3.25} & \textbf{84.78} \\
& Qwen2.5-32B & 52.42 & 33.36 & 48.69 & 19.13 & 18.92 & 76.95 & 77.30 & 76.99 & 58.57 & -4.12 & -3.43 & -3.31 & 84.68 \\

\Xhline{0.7pt}
\end{tabular}
\end{adjustbox}
\label{tab:csds_role_eval}
\end{table*}

\subsubsection{QMSum: Query-Oriented Long Dialogue Summarization}

Table~\ref{tab:subset_summarization_all} presents the performance of all evaluated models on the QMSum dataset, which features long-form meeting transcripts and evaluates both query-based and combined summarization scenarios. The experimental results reveal a consistent trend: non-reasoning LLMs achieve noticeably better performance than their reasoning-oriented counterparts across all domains and evaluation perspectives.

This advantage holds regardless of meeting type, whether academic, committee, or product-oriented discussions. In each case, the base models exhibit stronger alignment with human reference summaries, both in terms of lexical overlap and semantic adequacy. Notably, the superiority of base models is apparent not only in settings focused solely on query-driven summarization, but also when the evaluation involves a mix of query-based and generic summary objectives.

In contrast, reasoning LLMs do not deliver improvements in long-context, query-oriented summarization. Their performance is consistently lower than that of their base versions, regardless of the metric or specific meeting context. This pattern suggests that the additional instruction tuning for step-by-step reasoning does not benefit—and may even hinder—the ability of large language models to distill and organize salient information from lengthy and information-dense dialogue transcripts. One possible explanation is that such tuning encourages models to engage in unnecessary elaboration or over-reasoning, thereby reducing the focus and precision required for effective query-based summarization.

\begin{table*}[ht]
\centering
\caption{Summarization performance on the Academic, Committee, and Product domains, as well as the overall aggregate set (ALL) of \textsc{QMSum}, evaluated using ROUGE, BLEU, CHRF, BERTScore$_{F1}$, and COMET. For each metric, results are reported as \textit{specific} / \textit{all}, where \textit{specific} denotes query-based summarization, and \textit{all} denotes the aggregate performance over both query-based and generic summarization targets.}

\footnotesize
\begin{adjustbox}{max width=\textwidth}
\renewcommand{\arraystretch}{1.2}
\begin{tabular}{l lccccccc}
\Xhline{0.7pt}
\textbf{Datasets} & \textbf{Model} & \textbf{R1} & \textbf{R2} & \textbf{R-L} & \textbf{BLEU} & \textbf{CHRF} & \textbf{BERT$_{F1}$} & \textbf{COMET} \\
\Xhline{0.7pt}
\multirow{8}{*}{{Academic}}
& \textbf{Reasoning LLMs} & & & & & & & \\
& OpenAI-o1           & 22.53/23.17 & 4.24/4.12 & 13.52/13.37 & 1.13/1.07 & 28.55/30.32 & 84.46/84.22 & 63.97/64.38 \\
& DeepSeek-R1 & 19.80/20.80 & 3.98/3.79 & 12.32/12.32 & 1.16/1.09 & 26.54/28.94 & 83.67/83.53 & 62.34/62.78 \\
& QwQ-32B      & 16.79/17.85 & 3.59/3.52 & 10.54/10.62 & 0.84/0.81 & 22.86/25.12 & 82.67/82.55 & 62.10/62.31 \\
& \textbf{Non-Reasoning LLMs} & & & & & & & \\
& GPT-4o       & \textbf{24.26/25.00} & 5.23/5.14 & \textbf{15.51/15.45} & \textbf{1.74/1.66} & 30.67/32.66 & \textbf{85.51}/85.31 & 63.85/64.61 \\
& DeepSeek-V3  & 21.32/22.18 & \textbf{5.81/5.73} & 14.12/14.23 & 1.59/1.55 & 28.35/30.52 & 85.18/85.02 & \textbf{65.54/66.31} \\
& Qwen2.5-32B & 22.93/24.44 & 5.36/5.31 & 14.82/15.22 & 1.47/1.45 & \textbf{32.00/34.30} & 85.47/\textbf{85.49} & 64.32/65.04 \\
\hline
\multirow{8}{*}{{Committee}}
& \textbf{Reasoning LLMs} & & & & & & & \\
& OpenAI-o1           & 27.89/28.08 & 6.19/6.22 & 16.03/15.86 & 1.72/1.64 & 32.89/33.76 & 84.95/84.74 & 65.13/65.34 \\
& DeepSeek-R1 & 24.16/24.70 & 5.84/5.71 & 14.33/14.32 & 1.51/1.44 & 31.37/32.40 & 83.83/83.61 & 63.24/63.33 \\
& QwQ-32B      & 20.29/21.04 & 5.22/5.20 & 12.32/12.43 & 1.20/1.15 & 26.80/28.06 & 82.74/82.61 & 63.07/63.13 \\
& \textbf{Non-Reasoning LLMs} & & & & & & & \\
& GPT-4o       & \textbf{29.84/30.40} & 8.63/8.60 & \textbf{18.39/18.34} & \textbf{3.05/2.89} & 36.16/37.26 & 86.04/85.92 & 65.80/66.32\\
& DeepSeek-V3  & 27.12/27.57 & 8.53/8.50 & 17.07/17.05 & 2.71/2.57 & 34.25/35.27 & 85.51/85.38 & \textbf{67.00/67.41} \\
& Qwen2.5-32B & 29.03/29.76 & \textbf{8.87/8.80} & 17.89/18.00 & 2.96/2.85 & \textbf{36.56/37.64} & \textbf{85.92/85.87} & 66.71/67.19 \\
\hline
\multirow{8}{*}{{Product}}
& \textbf{Reasoning LLMs} & & & & & & & \\
& OpenAI-o1           & 29.54/29.89 & 6.35/6.57 & 16.51/16.44 & 1.79/1.69 & 33.62/34.94 & 85.26/85.06 & 65.76/66.09 \\
& DeepSeek-R1 & 25.62/26.17 & 6.10/6.02 & 15.02/15.01 & 1.52/1.43 & 32.81/33.77 & 84.03/83.74 & 63.97/64.03 \\
& QwQ-32B      & 21.97/22.85 & 5.72/5.73 & 13.18/13.34 & 1.29/1.23 & 28.86/30.12 & 82.91/82.77 & 64.01/64.11 \\
& \textbf{Non-Reasoning LLMs} & & & & & & & \\
& GPT-4o       & \textbf{30.80/31.55} & 8.85/8.97 & \textbf{18.59/18.63} & 2.77/2.62 & 36.93/38.15 & \textbf{86.24/86.17} & 67.18/67.77 \\
& Deepseek-V3  & 28.81/29.39 & 8.86/8.95 & 17.83/17.82 & 2.69/2.52 & 35.78/36.92 & 85.84/85.72 & \textbf{68.20/68.65} \\
& Qwen2.5-32B & 30.26/31.07 & \textbf{9.22/9.25} & 18.35/18.45 & \textbf{2.80/2.70} & \textbf{37.28/38.40} & 86.17/86.11 & 68.15/68.63 \\
\hline
\multirow{8}{*}{{ALL}}
& \textbf{Reasoning LLMs} & & & & & & & \\
& OpenAI-o1           & 27.89/28.08 & 6.19/6.22 & 16.03/15.86 & 1.72/1.64 & 32.89/33.76 & 84.95/84.74 & 65.13/65.34 \\
& DeepSeek-R1 & 24.16/24.70 & 5.84/5.71 & 14.33/14.32 & 1.51/1.44 & 31.37/32.40 & 83.83/83.61 & 63.24/63.33 \\
& QwQ-32B      & 20.29/21.04 & 5.22/5.20 & 12.32/12.43 & 1.20/1.15 & 26.80/28.06 & 82.74/82.61 & 63.07/63.13 \\
& \textbf{Non-Reasoning LLMs} & & & & & & & \\
& GPT-4o       & \textbf{29.84/30.40} & 8.63/8.60 & \textbf{18.39/18.34} & \textbf{3.05/2.89} & 36.16/37.26 & \textbf{86.04/85.92} & 65.80/66.32 \\
& DeepSeek-V3  & 27.12/27.57 & 8.53/8.50 & 17.07/17.05 & 2.71/2.57 & 34.25/35.27 & 85.51/85.38 & \textbf{67.00/67.41} \\
& Qwen2.5-32B & 29.03/29.76 & \textbf{8.87/8.80} & 17.89/18.00 & 2.96/2.85 & \textbf{36.56/37.64} & 85.92/85.87 & 66.71/67.19 \\
\Xhline{0.7pt}
\end{tabular}
\end{adjustbox}
\label{tab:subset_summarization_all}
\end{table*}

\section{Analysis}
In this section, we conduct further analyses to understand the intrinsic characteristics of generated summaries, assess summarization quality using LLM evaluators, examine explicit reasoning processes, and provide a detailed case study.
\subsection{Intrinsic Analysis of Summarization Quality}

\begin{table*}[htbp]
\centering
\caption{
Intrinsic quality metrics (mean $\pm$ std) of model-generated summarizations across four datasets. For CSDS, the \textit{final summary} is evaluated.}
\footnotesize
\begin{adjustbox}{max width=\textwidth}
\renewcommand{\arraystretch}{1.18}
\begin{tabular}{lrrrrrrrr}
\Xhline{0.7pt}
\textbf{Model} 
& \multicolumn{4}{c}{\textbf{SAMSum}} 
& \multicolumn{4}{c}{\textbf{DialogSum}} \\
\cmidrule(lr){2-5} \cmidrule(lr){6-9}
& \textbf{Length} & \textbf{Compression} & \textbf{Novelty} & \textbf{Coverage} 
& \textbf{Length} & \textbf{Compression} & \textbf{Novelty} & \textbf{Coverage} \\
\Xhline{0.7pt}
\multicolumn{9}{l}{\textbf{Reasoning LLMs}} \\
OpenAI-o1     & 28.84$\pm$11.54 & 3.11$\pm$1.86 & 0.68$\pm$0.13 & 0.34$\pm$0.14
              & 33.94$\pm$10.52 & 3.96$\pm$1.82 & 0.60$\pm$0.11 & 0.43$\pm$0.11 \\
DeepSeek-R1   & 29.75$\pm$11.92 & 2.97$\pm$1.63 & 0.70$\pm$0.12 & 0.32$\pm$0.13
              & 30.36$\pm$10.38 & 4.14$\pm$1.96 & 0.62$\pm$0.15 & 0.37$\pm$0.13 \\
QwQ-32B       & 27.39$\pm$14.46 & 3.39$\pm$2.44 & 0.69$\pm$0.13 & 0.33$\pm$0.13
              & 29.80$\pm$21.12 & 4.88$\pm$2.78 & 0.66$\pm$0.12 & 0.37$\pm$0.13 \\
\multicolumn{9}{l}{\textbf{Non-Reasoning LLMs}} \\
GPT-4o        & 27.95$\pm$12.07 & 3.19$\pm$1.73 & 0.65$\pm$0.13 & 0.37$\pm$0.14
              & 31.24$\pm$9.86  & 4.32$\pm$2.00 & 0.59$\pm$0.12 & 0.44$\pm$0.12 \\
DeepSeek-V3   & 28.21$\pm$11.99 & 3.18$\pm$1.78 & 0.66$\pm$0.13 & 0.36$\pm$0.14
              & 29.97$\pm$9.61  & 4.51$\pm$2.07 & 0.58$\pm$0.11 & 0.44$\pm$0.12 \\
Qwen2.5-32B   & 30.47$\pm$13.07 & 2.95$\pm$1.64 & 0.67$\pm$0.13 & 0.35$\pm$0.14
              & 31.09$\pm$10.17 & 4.42$\pm$2.17 & 0.61$\pm$0.12 & 0.42$\pm$0.12 \\
\hline
Reference summary & 20.02$\pm$10.65 & 5.05$\pm$3.73 & 0.60$\pm$0.16 & 0.41$\pm$0.16
              & 18.75$\pm$8.65  & 7.26$\pm$2.36 & 0.56$\pm$0.16 & 0.46$\pm$0.16  \\
\Xhline{0.7pt}
\textbf{Model} 
& \multicolumn{4}{c}{\textbf{CSDS}} 
& \multicolumn{4}{c}{\textbf{QMSum}} \\
\cmidrule(lr){2-5} \cmidrule(lr){6-9}
& \textbf{Length} & \textbf{Compression} & \textbf{Novelty} & \textbf{Coverage} 
& \textbf{Length} & \textbf{Compression} & \textbf{Novelty} & \textbf{Coverage} \\
\Xhline{0.7pt}
\multicolumn{9}{l}{\textbf{Reasoning LLMs}} \\
OpenAI-o1     & 39.11$\pm$12.10 & 12.92$\pm$5.10 & 0.48$\pm$0.10 & 0.57$\pm$0.10
              & 194.78$\pm$137.90 & 72.76$\pm$53.13 & 0.48$\pm$0.11 & 0.61$\pm$0.10 \\
DeepSeek-R1   & 38.03$\pm$12.09 & 13.02$\pm$5.53 & 0.49$\pm$0.11 & 0.54$\pm$0.12
              & 206.58$\pm$81.91 & 53.86$\pm$35.88 & 0.59$\pm$0.11 & 0.52$\pm$0.10 \\
QwQ-32B       & 42.08$\pm$13.10 & 11.83$\pm$4.31 & 0.52$\pm$0.10 & 0.52$\pm$0.10
              & 312.45$\pm$123.44 & 35.67$\pm$24.68 & 0.61$\pm$0.10 & 0.52$\pm$0.10 \\
\multicolumn{9}{l}{\textbf{Non-Reasoning LLMs}} \\
GPT-4o        & 36.12$\pm$10.33 & 13.70$\pm$5.52 & 0.45$\pm$0.11 & 0.57$\pm$0.11
              & 161.80$\pm$85.78 & 84.86$\pm$94.12 & 0.41$\pm$0.13 & 0.69$\pm$0.10 \\
DeepSeek-V3   & 34.46$\pm$10.48 & 14.62$\pm$5.89 & 0.42$\pm$0.11 & 0.61$\pm$0.10
              & 225.15$\pm$111.68 & 53.39$\pm$37.41 & 0.45$\pm$0.12 & 0.68$\pm$0.09 \\
Qwen2.5-32B   & 37.89$\pm$11.69 & 13.26$\pm$5.07 & 0.46$\pm$0.10 & 0.57$\pm$0.10
              & 176.19$\pm$77.60 & 65.28$\pm$42.57 & 0.47$\pm$0.13 & 0.66$\pm$0.10 \\
\hline
Reference summary & 46.57$\pm$27.24 & 12.67$\pm$6.68 & 0.38$\pm$0.13 & 0.60$\pm$0.12
              & 64.72$\pm$29.22 & 184.13$\pm$138.82 & 0.27$\pm$0.12 & 0.77$\pm$0.10 \\
\Xhline{0.7pt}
\label{tab:length_novelty_coverage}
\end{tabular}
\end{adjustbox}
\end{table*}

For outputs on all four datasets, we evaluate three intrinsic summary qualities: \textit{Length}, \textit{compression rate}~\citep{grusky-etal-2018-newsroom}, and \textit{abstractiveness}~\citep{bommasani_intrinsic_2020}. Specifically, the compression rate measures the length difference between the input dialogue and the summary. To assess abstractiveness, we consider two complementary aspects: \textit{Extractive Fragment Coverage}~\citep{grusky-etal-2018-newsroom}, which quantifies the proportion of the summary that can be directly “copied” from the input text, and \textit{Novelty}~\citep{liu_improving_2023}, which calculates the ratio of words in the summary that do not appear in the input dialogue.

\textit{Compression rate} is computed as the ratio of the number of words in the input dialogue $D$ to those in the summary~$Y$:
\begin{equation}
    \mathrm{Compression}(D, Y) = \frac{|D|}{|Y|}.
\end{equation}

\textit{Extractive fragment coverage} measures the extent to which the summary can be directly copied from the input, defined as the proportion of summary tokens that belong to extractive fragments shared with the input text:
\begin{equation}
    \mathrm{Coverage}(D, Y) = \frac{1}{|Y|} \sum_{f \in F(D, Y)} |f|,
\end{equation}
where $F(D, S)$ is the set of word sequences that are common between $D$ and $S$.

\textit{Novelty} quantifies the degree of abstraction by calculating the proportion of summary tokens that do not appear in the input:
\begin{equation}
    \mathrm{Novelty}(D, Y) = 1 - \frac{|D \cap Y|}{|Y|},
\end{equation}
where $|D \cap Y|$ denotes the number of words shared between the summary and the input.

Table~\ref{tab:length_novelty_coverage} summarizes the intrinsic evaluation results across four datasets. A consistent observation for English datasets (SAMSum, DialogSum, QMSum) is that reasoning LLMs generally produce longer summaries, resulting in lower compression rates compared to their non-reasoning counterparts. Specifically, for SAMSum and DialogSum, reasoning LLM-generated summaries exhibit higher novelty but lower coverage scores, indicating these summaries tend to include more abstractive content at the cost of potentially reduced fidelity to the original dialogue. In contrast, non-reasoning LLMs generate summaries closer in length and extractiveness to reference summaries, achieving higher coverage scores, suggesting better alignment with human summarization styles.

However, the Chinese dataset (CSDS) reveals a notably different pattern. Non-reasoning LLMs generate more concise summaries (shorter length and higher compression rates) and achieve higher coverage compared to reasoning LLMs. Meanwhile, reasoning LLMs still show higher novelty, consistent with the English datasets, but this increased abstraction does not translate into improved summary quality or fidelity. Interestingly, the reference summaries for CSDS are longer and exhibit similar compression rates to model-generated summaries, reflecting that human-written summaries in this domain often contain detailed, factual descriptions rather than highly compressed abstractions.

Across all datasets, human-edited reference summaries consistently demonstrate the highest coverage, emphasizing their superior effectiveness in faithfully aggregating essential dialogue content. Moreover, reference summaries generally exhibit lower novelty, reflecting careful human editing aimed at factual accuracy and content fidelity, particularly notable in QMSum and CSDS datasets that involve longer and more complex dialogue contexts.

\subsection{Comparative Assessment of Summarization Quality with LLM Evaluators}

\begin{figure*}
    \centering
    \includegraphics[width=\textwidth]{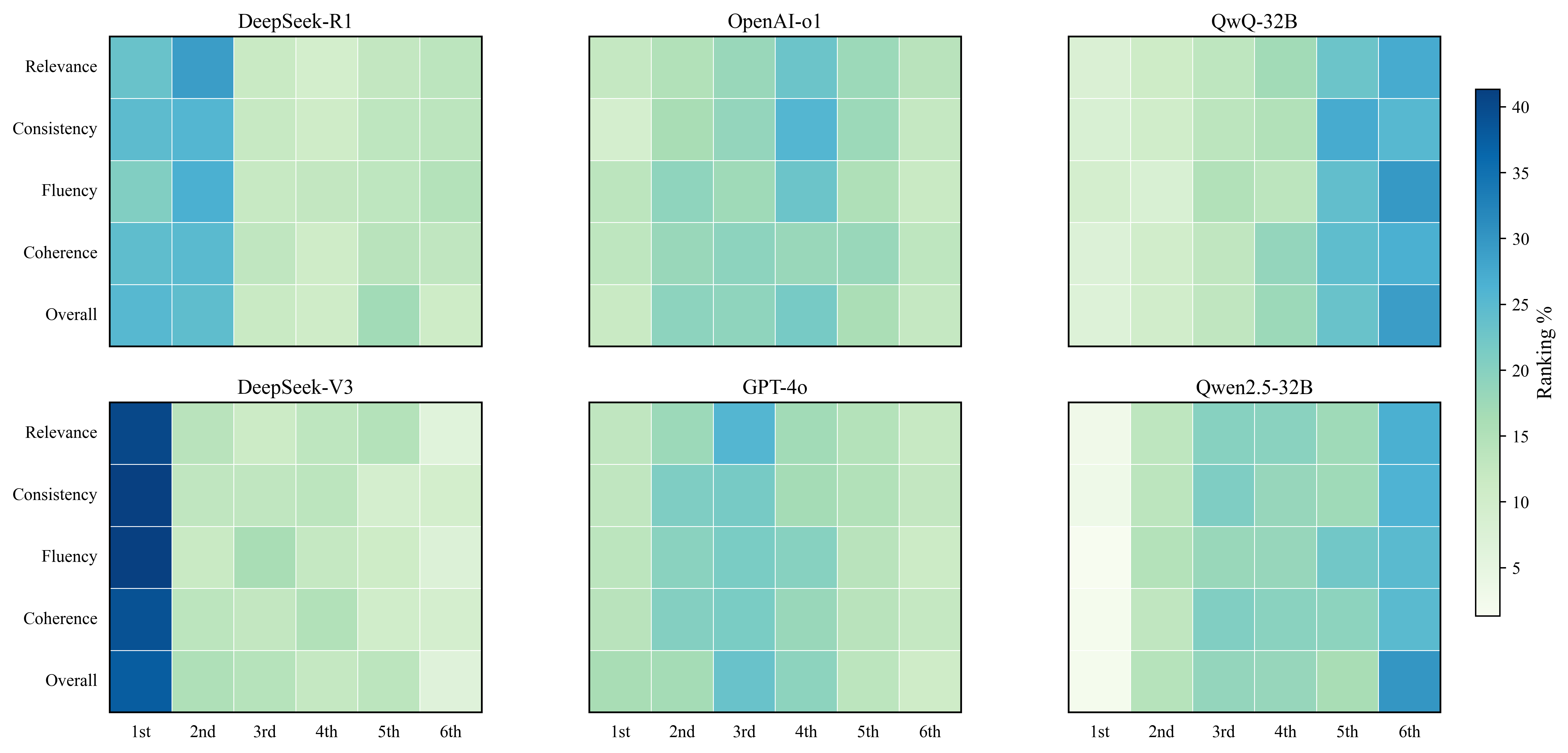}
    \caption{Distribution of Model Rankings across Evaluation Dimensions. 
The heatmap illustrates the percentage distribution of each model’s rank (1st–6th) for all five evaluation criteria, as assessed by LLM judges.}
    \label{heatmap}
\end{figure*}
Recent studies have demonstrated that LLMs can effectively serve as evaluators for natural language generation tasks, showing strong alignment with human judgments across various evaluation dimensions~\citep{gao2025llm,murugadoss2025evaluating,gu2024survey}. Inspired by these findings, we complemented the computation of the summarization metrics with an LLM-based evaluation strategy to further examine the summarization quality of reasoning-oriented and non-reasoning models. Specifically, we selected three non-reasoning LLMs (DeepSeek-V3, GPT-4o, and Qwen2.5-32B) as evaluators to assess summaries generated by three reasoning-oriented models (DeepSeek-R1, OpenAI-o1, and QwQ-32B) and their corresponding non-reasoning counterparts.

We randomly sampled 100 dialogue-summary pairs from the SAMSum dataset and instructed the evaluator models to rank candidate summaries according to their alignment with the ground-truth reference summary. Specifically, the evaluators assessed summaries based on five evaluation dimensions adapted from prior human evaluation studies~\citep{kryscinski2019neural}. The prompt used for evaluation is detailed in Appendix~\ref{fig:prompt_ranking}.

\textbf{Relevance} Whether the summary effectively captures the key points and central content of the original dialogue.

\textbf{Consistency} Whether the summary accurately reflects information presented in the dialogue without introducing factual errors or unsupported details.

\textbf{Fluency} Whether the summary is grammatically correct, natural, and easy to read. 

\textbf{Coherence} Whether the summary is logically organized and clearly structured. 

\textbf{Overall} A comprehensive judgment considering all of the above aspects. 

The evaluator models were explicitly instructed to rank candidate summaries from best to worst, with the flexibility to group summaries as equally good when distinctions in quality were subtle or negligible. Allowing equal rankings mitigates the bias introduced by forced distinctions and better aligns with realistic human evaluation practices.
To verify the reliability and consistency of the LLM-based evaluation results, we measured the inter-annotator agreement across evaluator models using Krippendorff's alpha~\citep{hayes2007answering}. The resulting alpha coefficients indicated satisfactory levels of agreement for all five evaluation dimensions, specifically: \textit{Relevance} ($\alpha=0.553$), \textit{Consistency} ($\alpha=0.562$), \textit{Fluency} ($\alpha=0.554$), \textit{Coherence} ($\alpha=0.583$), and \textit{Overall} ($\alpha=0.562$).

Figure~\ref{heatmap} presents the distribution of rankings for each model across the five evaluation dimensions. From the heatmap, it is clearly observed that among all evaluated models, DeepSeek-V3 consistently achieved the highest proportion of top rankings, particularly evident in the dimensions of \textit{Relevance}, \textit{Consistency}, and \textit{Overall}. Additionally, when directly comparing reasoning-oriented models with their non-reasoning counterparts, the non-reasoning models consistently demonstrated superior performance across all evaluation dimensions, underscoring that explicit reasoning optimization does not necessarily translate into improved summarization quality for general dialogues.

\begin{figure*}
    \centering
    \includegraphics[width=\textwidth]{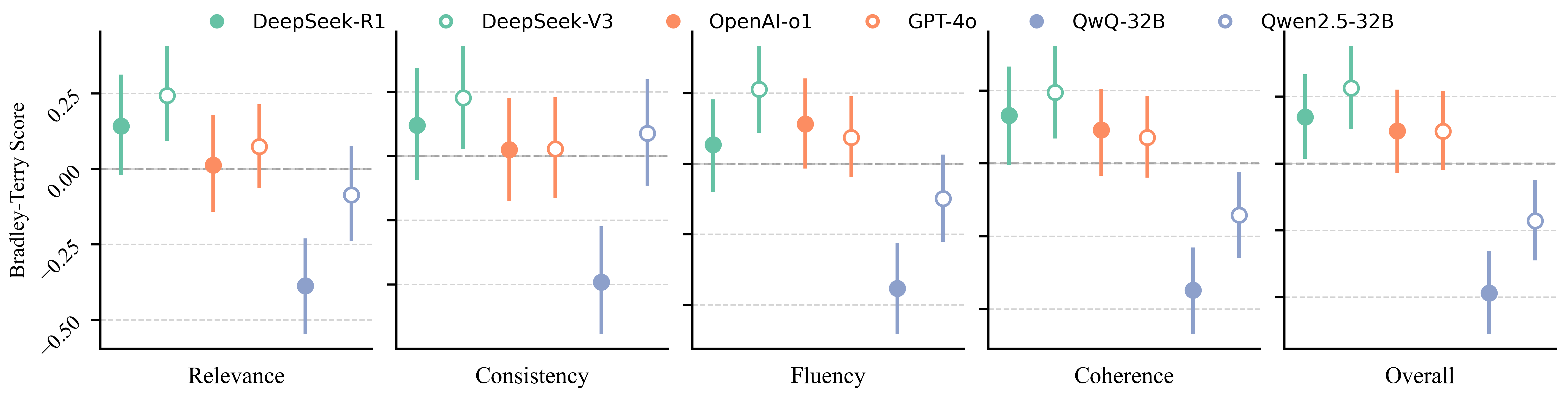}
    \caption{Bradley-Terry scores with 95\% confidence intervals across five evaluation dimensions.}
    \label{btscore}
\end{figure*}

To enable rigorous statistical comparison across models based on the collected ranking data, we further analyzed the results using the Bradley–Terry (BT) model~\citep{bradley1952rank}. The BT model is specifically designed to estimate latent quality scores from pairwise preferences derived from ranking assignments, providing a principled way to summarize overall model performance across evaluators and samples.

Concretely, each ranking of candidate summaries provided by the evaluator models can be interpreted as a sequence of pairwise wins and losses among models. For any two models $i$ and $j$, if $i$ is ranked higher than $j$, this is treated as a pairwise win for $i$. The BT model then defines the probability of model $i$ being preferred over model $j$ as:
\begin{equation}
    P(i \succ j) = \frac{\exp(s_i)}{\exp(s_i) + \exp(s_j)},
\end{equation}
where $s_i$ and $s_j$ denote the latent (log-scale) quality scores for models $i$ and $j$, respectively. The set of scores $\{s_i\}$ is estimated by maximizing the likelihood of all observed pairwise outcomes derived from the evaluator rankings. To further account for sampling variability and to obtain uncertainty estimates, we employed 1{,}000 bootstrap resamplings when fitting the BT model, thereby reporting the mean and 95\% confidence intervals for each model's score in every evaluation dimension.

Figure~\ref{btscore} presents the Bradley–Terry scores across five evaluation dimensions: relevance, consistency, fluency, coherence, and overall summary quality. The results show clear trends aligning well with our earlier automatic evaluation findings. Specifically, DeepSeek-V3 consistently outperforms its reasoning-oriented counterpart DeepSeek-R1 across all evaluation dimensions, demonstrating strong capabilities in generating summaries that are more relevant, consistent, fluent, and coherent. Similarly, Qwen2.5-32B achieves notably higher scores compared to its reasoning counterpart QwQ-32B, indicating that reasoning-specific training did not yield improvements in summary quality for this model family. Regarding OpenAI-o1 and GPT-4o, the performance differences are subtler: GPT-4o slightly surpasses OpenAI-o1 on relevance but exhibits comparable performance on consistency, fluency, coherence, and overall quality dimensions. Notably, both OpenAI-o1 and GPT-4o generally outperform Qwen-family models (Qwen2.5-32B and QwQ-32B) across most metrics, except for fluency, where differences are minimal.

To better understand the relationship between automatic and LLM-based evaluation, we also report the correlation between these metrics in Appendix~\ref{fig:autoLLM_corr}. This analysis reveals that the correlation between automatic metrics and LLM-based evaluation remains generally low, consistent with previous findings in other NLG tasks~\citep{reiter_structured_2018, elangovan2025beyond}.

\subsection{In-depth Analysis of Reasoning Processes}

To gain deeper insight into model reasoning behavior, we conducted a detailed analysis of the explicit reasoning processes produced by DeepSeek-R1 for the same set of 100 randomly sampled dialogue instances from the SAMSum dataset described above. Our evaluation framework draws on recent advances in the assessment of step-by-step reasoning traces~\citep{lee2025evaluating}, and we extend this framework by introducing an additional criterion, \textit{depth}, which specifically evaluates the presence of abstraction, synthesis, and integration within the reasoning process. Prior work has demonstrated that explicit inductive and summarization steps during the reasoning process can enhance the conciseness and informativeness of generated summaries~\citep{sun2024prompt}. Each dimension in our evaluation is designed to reflect a distinct functional property of reasoning quality:

\textbf{Relevance} Assesses whether each reasoning step is directly grounded in the source dialogue, avoiding hallucinated, irrelevant, or unsupported content.

\textbf{Validity} Measures the logical correctness of individual reasoning steps, ensuring that every inference is entailed by either the dialogue or previous steps, without contradictions or faulty logic.

\textbf{Coherence} Evaluates the semantic and structural connectivity of the reasoning chain, examining whether all steps are linked smoothly and the overall chain progresses without abrupt jumps, redundancy, or gaps.

\textbf{Utility} Quantifies the informativeness and necessity of each reasoning step, determining whether each contributes meaningfully toward constructing a correct summary or answer, rather than being redundant or trivial.

\textbf{Depth} Examines whether the reasoning chain demonstrates abstraction, integration, or higher-level inference—moving beyond surface-level repetition or paraphrase to capture more complex relationships within the dialogue.

For each evaluation dimension, we designed a prompt that instructed the evaluator to assign an integer score from 1 (very poor) to 5 (excellent), based solely on the specified aspect of the reasoning process.  The prompt used for evaluation is detailed in Appendix~\ref{fig:prompt_reasoning_scoring}. The scoring was conducted independently by three advanced LLMs—DeepSeek-V3, GPT-4o, and GPT-4.1—serving as automatic judges. This setup enables a granular assessment of the strengths and weaknesses of model-generated reasoning, revealing how well each approach performs across multiple facets of process-level reasoning in dialogue summarization tasks.

\begin{figure*}
   \centering
   \includegraphics[width=0.9\textwidth]{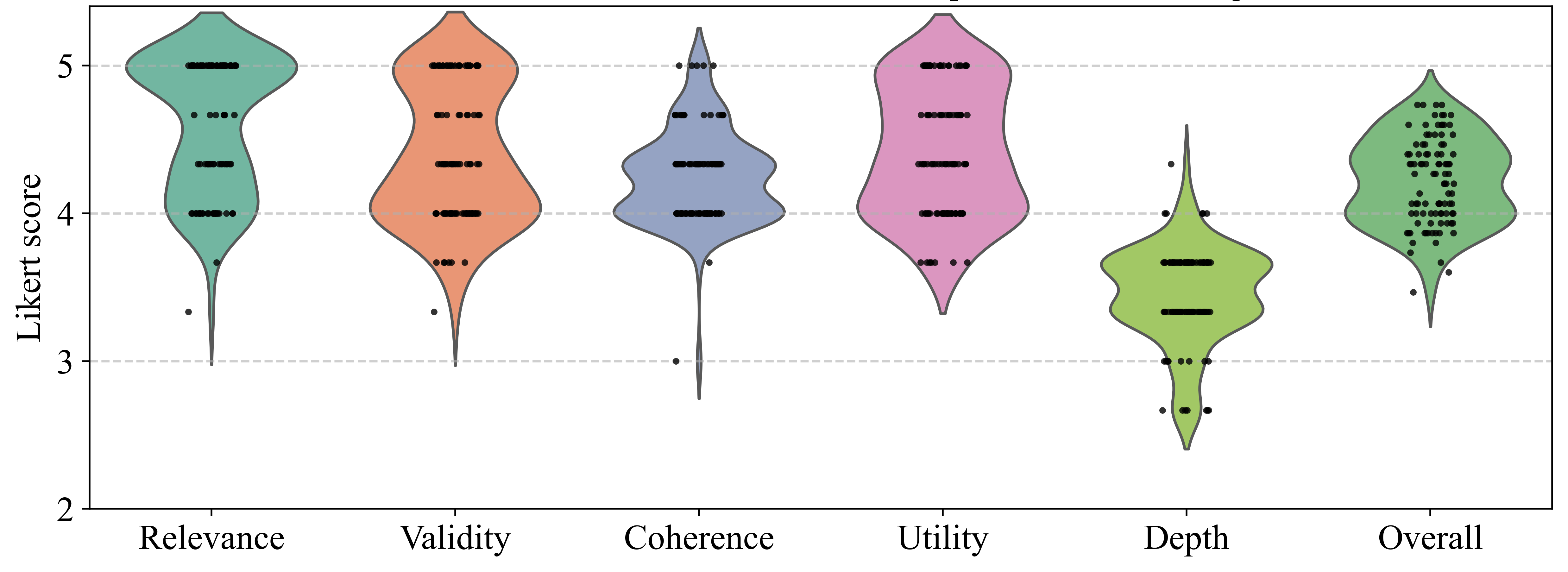}
   \caption{Distribution of \textit{DeepSeek-R1} Reasoning Processes across Evaluation Dimensions.}
   \label{tiqin}
\end{figure*}

Figure~\ref{tiqin} shows the distribution of scores assigned by LLM evaluators across the five specific evaluation dimensions (\textit{Relevance}, \textit{Validity}, \textit{Coherence}, \textit{Utility}, and \textit{Depth}) as well as an aggregated \textit{Overall} dimension, for the reasoning processes generated by \textit{DeepSeek-R1}. We observe that the model achieves generally high ratings in \textit{Relevance} and \textit{Validity}, with most scores clustering around 4 or 5, indicating strong factual grounding and logical correctness in reasoning steps. Conversely, scores for \textit{Depth} and \textit{Utility} exhibit broader distributions and notably lower medians, suggesting variability in the model’s capacity for higher-level abstraction and efficient information synthesis. Additionally, the distribution for \textit{Coherence} reveals moderate consistency, implying occasional difficulties in maintaining smooth logical transitions between reasoning steps. The aggregated \textit{Overall} scores demonstrate a relatively centralized distribution, reflecting balanced performance across these dimensions but also highlighting room for improvement, particularly regarding deeper, integrative reasoning abilities.

Notably, the relatively lower scores observed in the \textit{Depth} dimension indicate that the explicit reasoning processes generated by DeepSeek-R1 often lack sufficient abstraction, integration, and higher-level summarization capabilities. This finding suggests a potential reason behind the observed performance gap, where reasoning-oriented models fail to outperform their non-reasoning counterparts in dialogue summarization tasks. Specifically, the limited depth in reasoning processes may restrict these models from effectively synthesizing information and capturing nuanced dialogue content, ultimately constraining their summarization quality.
\begin{figure*}
   \centering
   \includegraphics[width=0.95\textwidth]{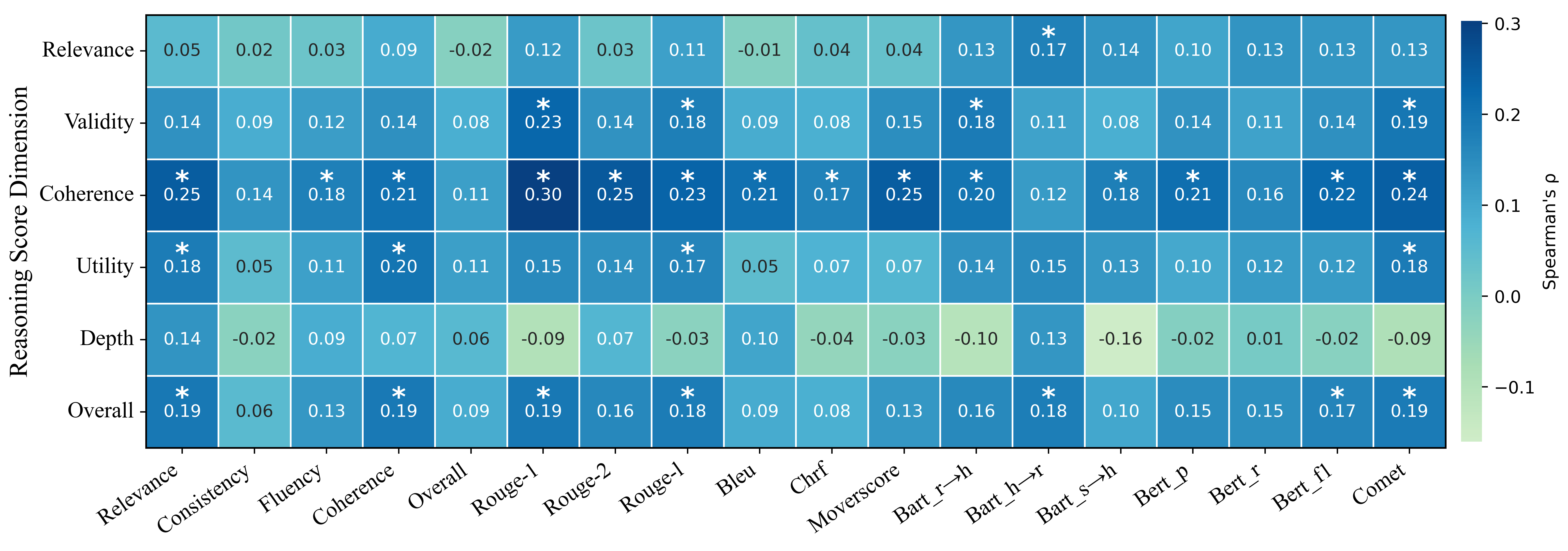}
   \caption{
Spearman correlations between reasoning process evaluation dimensions and summarization quality metrics (LLMs evaluation: Relevance--Overall; automatic metrics: ROUGE-1--COMET). Asterisks (*) denote statistically significant correlations ($p<0.05$).
}
    \label{fig:reasoning_summary_correlation}

\end{figure*}

Figure~\ref{fig:reasoning_summary_correlation} illustrates the correlation between reasoning process scores across evaluation dimensions and various summarization quality metrics. Contrary to our initial hypothesis—that higher-quality reasoning processes, particularly in dimensions like \textit{Depth} and \textit{Coherence}, would positively correlate with summarization performance—our analysis reveals predominantly weak and statistically insignificant correlations. This indicates a possible disconnect between explicitly assessed reasoning quality and summarization effectiveness.

\subsection{Case Study}

Table~\ref{tab:case_study_r1_vs_v3} presents a representative example comparing DeepSeek-R1 (reasoning) and DeepSeek-V3 (non-reasoning) on a dialogue summarization task. In this case, the ground truth summary is concise, directly highlighting the key facts: the accident location and the absence of fatalities. DeepSeek-V3 closely matches this ground truth, providing a short and accurate summary.

In contrast, DeepSeek-R1’s generated summary is noticeably more verbose and includes details that are not central to the main event, such as identifying all participants and their emotional reactions. Examination of DeepSeek-R1’s explicit reasoning process reveals several key phenomena. First, a significant portion of the reasoning steps consists of direct paraphrasing or repetition of the dialogue, rather than abstraction or synthesis. These repetitive segments introduce redundancy and increase the risk of introducing noisy or irrelevant information. Second, the reasoning process introduces hallucinated or factually inconsistent content (e.g., “I tried to get to the Circle Mall” is highlighted as a possible hallucination when not directly corroborated by the context), which is subsequently propagated to the generated summary.

Additionally, the stepwise reasoning process attempts to consider and weigh multiple aspects and reactions in the dialogue, but this breadth of consideration comes at the expense of clarity and conciseness. Instead of distilling the dialogue down to its most salient facts, the model’s reasoning leads to summaries that are overly detailed and less focused. This example demonstrates that, in the context of dialogue summarization, the inclusion of explicit multi-step reasoning does not inherently improve—and may even harm—summary quality, by increasing both verbosity and the risk of factual errors or inconsistencies.

\begin{table*}[htbp]
\centering
\caption{
Case study comparing DeepSeek-R1 and DeepSeek-V3 on a representative dialogue summarization example. In the reasoning process column, \textcolor{red}{red text} indicates templated or redundant sentences, \hlyellow{yellow highlights} mark hallucinated or factually inconsistent content, and \hlblue{blue text} identifies the model’s actual abstraction or synthesis steps within the reasoning process. }
\footnotesize
\begin{adjustbox}{max width=\textwidth}
\renewcommand{\arraystretch}{1.2}
\begin{tabular}{p{0.14\textwidth} | p{0.81\textwidth}}
\Xhline{0.7pt}
\textbf{Source} & \textbf{Content} \\
\Xhline{0.7pt}
Dialogue & \makecell[l]{Oli: Theres a car accident \\
Katie: Where? \\
Oli: On circle drive \\
Oli: I tried to \hlyellow{get to the Circle Mall} \\
Katie: Oh no \\
Pavel: Its on the news now \\
Pavel: Theres no deaths \\
Katie: Thank god \textit{(angel emoji)}} \\
\hline
Ground Truth & There has been an accident on Circle Drive, near Circle Mall. There are no fatalities. \\
\hline
DeepSeek-V3 & There was a car accident on Circle Drive, but thankfully there were no deaths. \\
\hline
DeepSeek-R1 & Oli informs Katie and Pavel about a car accident on \hlyellow{Circle Drive near Circle Mall}. Pavel confirms it's on the news with no fatalities, and Katie expresses relief. \\
\hline
\makecell[l]{DeepSeek-R1\\Reasoning Process} & 
\textcolor{red}{Okay, let me try to figure out how to summarize this dialogue. First, I need to read through the conversation carefully to understand the main points.} \\[0.2em]
& So, Oli starts by saying there's a car accident. Katie asks where, and Oli responds it's on Circle Drive, and he tried to get to Circle Mall. Then Pavel mentions it's on the news now and that there are no deaths. Katie is relieved and says "Thank god" with an emoji. \\[0.2em]
& \textcolor{red}{Looking at the examples given, summaries capture the key actions and outcomes without extra details. For instance, in the previous summaries, they mention who is involved and the main event or conclusion.} \\[0.2em]
& Here, the main points are: \hlblue{a car accident occurred \hlyellow{on Circle Drive, near Circle Mall}. Pavel reports it's on the news with no fatalities, and Katie is relieved.}The key elements are the accident location, no deaths, and the reactions. \\[0.2em]
& I should make sure the \hlblue{summary includes the accident location, that there were no deaths, and maybe Katie's relief. Also, Oli tried to go to the mall, which might explain why he's mentioning the accident. But the main focus is the accident itself and the outcome.} So the summary should be something like: Oli reports a car accident on \hlyellow{Circle Drive near Circle Mall}. Pavel says there are no deaths, and Katie is relieved. \\
\Xhline{0.7pt}
\end{tabular}
\end{adjustbox}
\label{tab:case_study_r1_vs_v3}
\end{table*}

\section{Conclusion}

In this work, we present the first systematic, scenario-driven comparison of reasoning LLMs and their conventional counterparts for dialogue summarization. By evaluating strong models—including OpenAI-o1, DeepSeek-R1, QwQ-32B, and matched base models—across generic, role-oriented, and query-oriented paradigms, we provide a detailed assessment of current capabilities and limitations.

Our results demonstrate that, contrary to the gains observed in other reasoning-intensive domains~\citep{chen_evaluating_2025}, explicit stepwise reasoning does not necessarily confer an advantage for dialogue summarization. Instead, reasoning models often fall short of strong base LLMs like GPT-4o, DeepSeek-V3, and Qwen2.5-32B, with common issues including verbosity, factual inconsistencies, and loss of conciseness—especially in tasks that require high-level abstraction.

Through a multi-perspective evaluation combining automatic metrics, LLM-based judgment, and scenario-specific error analysis, we further reveal that current benchmarks and evaluators still struggle to fully capture the challenges of dialogue summarization. Although LLM-based evaluation offers better alignment with human judgment than traditional metrics, it remains difficult to robustly assess factuality, conciseness, and pragmatic adequacy.

Our analysis underscores the importance of developing more nuanced modeling and evaluation strategies that effectively balance stepwise reasoning, abstraction, and faithfulness in dialogue summarization. We believe that our comprehensive evaluation and findings will provide useful guidance for future research and help inform the design and assessment of LLMs in real-world dialogue summarization tasks.

\bibliographystyle{cas-model2-names}
\bibliography{cas-refs}

\appendix
\renewcommand{\thefigure}{\Alph{section}.\arabic{figure}}
\setcounter{figure}{0}

\section{Prompt Template}

\begin{figure}[htbp]
\centering
\caption{Prompt Template for LLM-based Reasoning Process Scoring.}
\includegraphics[width=0.8\textwidth]{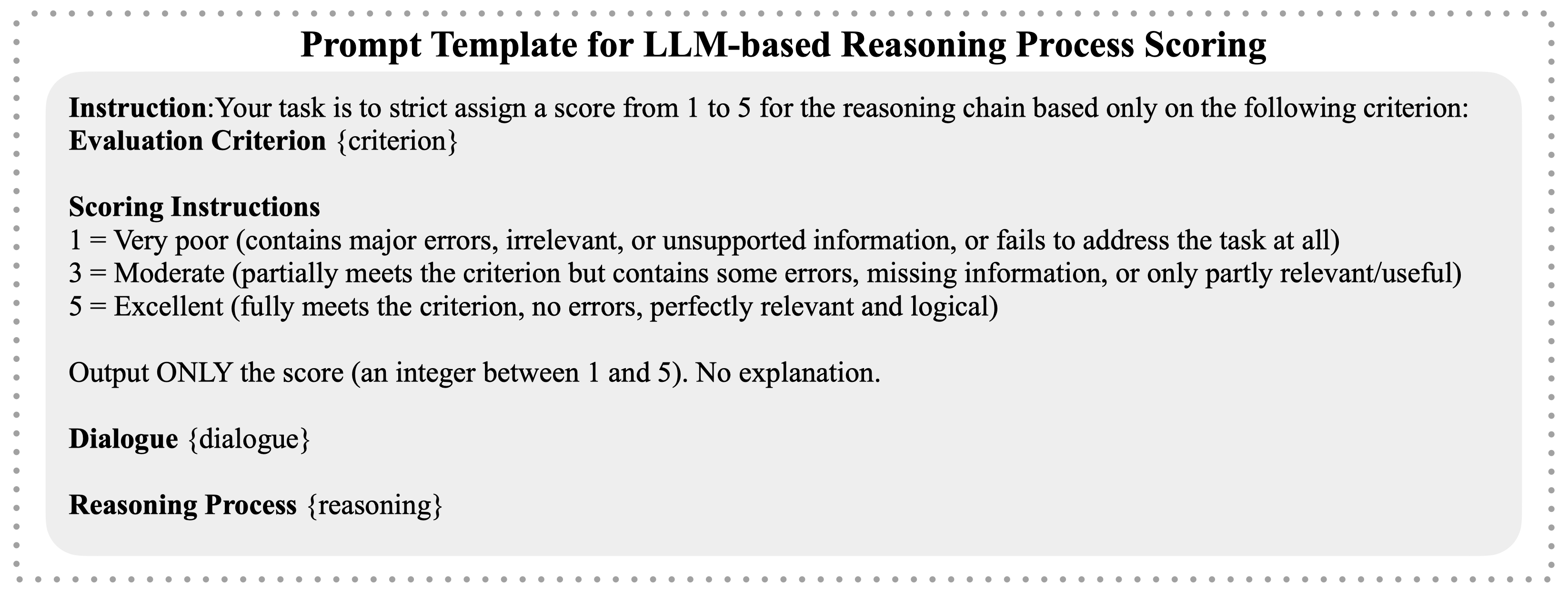}
\label{fig:prompt_reasoning_scoring}
\end{figure}

\begin{figure}[htbp]
\centering
\caption{Prompt Template for LLM-based Ranking Evaluation.}
\includegraphics[width=0.8\textwidth]{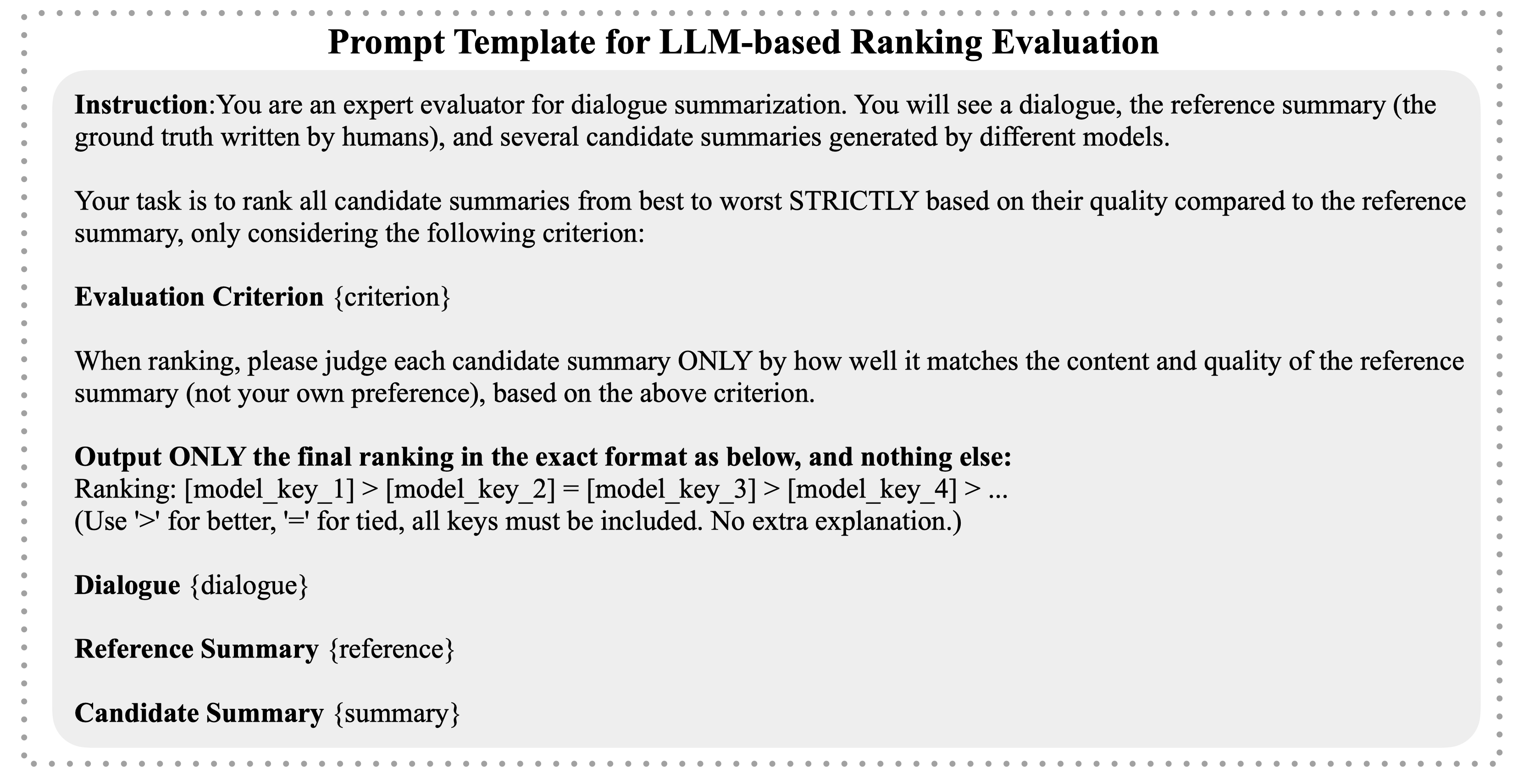}
\label{fig:prompt_ranking}
\end{figure}

\section{Correlations between automatic evaluation and LLMs evaluation}
Figure~\ref{fig:autoLLM_corr} shows that the correlations between automatic evaluation metrics and LLM-based evaluation in dialogue summarization are generally low. This observation is consistent with previous studies in other NLG tasks~\citep{reiter_structured_2018,elangovan2025beyond}, Together, these findings highlight the limitations of automatic metrics for evaluating summary quality, reinforcing the need for more human-aligned assessment approaches.
\setcounter{figure}{0} 
\begin{figure}[htbp]
\centering
\caption{The correlation (Pearson’s $r$) between different automatic evaluation metrics and LLMs evaluation.}
\includegraphics[width=0.73\textwidth]{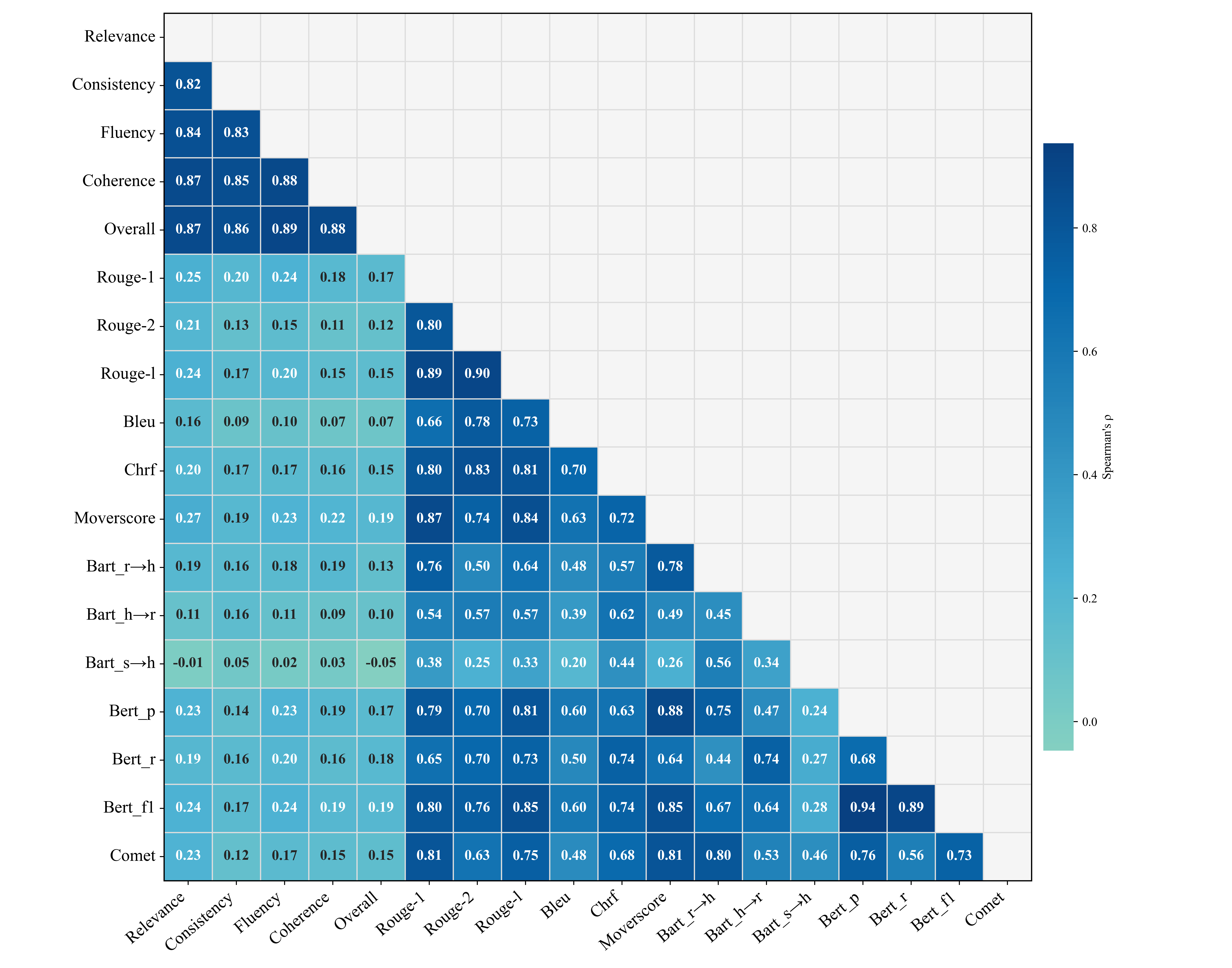}
\label{fig:autoLLM_corr}
\end{figure}

\end{document}